\documentclass{article}

\usepackage[final,nonatbib]{neurips_2021}

\usepackage[utf8]{inputenc} 
\usepackage[T1]{fontenc}    
\usepackage[numbers,sort&compress]{natbib}
\usepackage[pdftex,%
	pdfauthor={Gerrit J.J. van den Burg and Christopher K.I. Williams},%
	pdftitle={On Memorization in Probabilistic Deep Generative Models},%
	pdfkeywords={Memorization, Generative Models, Density Estimation, Unsupervised Learning},%
	pdfcreator={},%
	pdfproducer={},%
	pdfcreationdate={},%
	hidelinks]{hyperref}        
\usepackage{url}            
\usepackage{booktabs}       
\usepackage{amsfonts}       
\usepackage{nicefrac}       
\usepackage{microtype}      
\usepackage[dvipsnames]{xcolor}         
\usepackage{graphicx}
\usepackage{wrapfig}
\usepackage{placeins}

\usepackage{amsmath}
\usepackage{amsfonts}
\usepackage{amssymb}
\usepackage{subcaption}

\usepackage{algorithm}
\usepackage{algpseudocode}

\newcommand{\bft}[1]{\mathbf{#1}}
\newcommand{\bfs}[1]{\boldsymbol{#1}}

\newcommand{\abs}[1]{\left|#1\right|}
\newcommand{\myd}{\,\mathrm{d}}
\newcommand{\given}{\,|\,}

\definecolor{MyBlue}{HTML}{004488}
\definecolor{MyYellow}{HTML}{DDAA33}
\definecolor{MyRed}{HTML}{BB5566}

\DeclareRobustCommand{\bbone}{\text{\usefont{U}{bbold}{m}{n}1}}

\title{On Memorization in Probabilistic\\ Deep Generative Models}

\author{%
	Gerrit J.J.~van~den~Burg\thanks{Work done while at The Alan Turing 
		Institute.} \\
	\texttt{gertjanvandenburg@gmail.com} \\
	\And
	Christopher K.I.~Williams \\
	University of Edinburgh \\
	The Alan Turing Institute \\
	\texttt{ckiw@inf.ed.ac.uk}
}

\begin{document}

\maketitle

\begin{abstract}
	Recent advances in deep generative models have led to impressive 
	results in a variety of application domains. Motivated by the 
	possibility that deep learning models might memorize part of the input 
	data, there have been increased efforts to understand how memorization 
	arises. In this work, we extend a recently proposed measure of 
	memorization for supervised learning (Feldman, 2019) to the 
	unsupervised density estimation problem and adapt it to be more 
	computationally efficient. Next, we present a study that demonstrates 
	how memorization can occur in probabilistic deep generative models 
	such as variational autoencoders. This reveals that the form of 
	memorization to which these models are susceptible differs 
	fundamentally from mode collapse and overfitting. Furthermore, we show 
	that the proposed memorization score measures a phenomenon that is not 
	captured by commonly-used nearest neighbor tests. Finally, we discuss 
	several strategies that can be used to limit memorization in practice.  
	Our work thus provides a framework for understanding problematic 
	memorization in probabilistic generative models.
\end{abstract}

\section{Introduction}
\label{sec:intro}
In the last few years there have been incredible successes in generative 
modeling through the development of deep learning techniques such as 
variational autoencoders (VAEs) \cite{kingma2014auto,rezende2014stochastic}, 
generative adversarial networks (GANs) \cite{goodfellow2014generative}, 
normalizing flows \cite{tabak2013family,rezende2015variational}, and diffusion 
networks \cite{sohl2015deep,ho2020denoising}, among others. The goal of 
generative modeling is to learn the data distribution of a given data set, 
which has numerous applications such as creating realistic synthetic data, 
correcting data corruption, and detecting anomalies. Novel architectures for 
generative modeling are typically evaluated on how well a complex, high 
dimensional data distribution can be learned by the model and how realistic 
the samples from the model are. An important question in the evaluation of 
generative models is to what extent training observations are \emph{memorized} 
by the learning algorithm, as this has implications for data privacy, model 
stability, and generalization performance. For example, in a medical setting 
it is highly desirable to know if a synthetic data model could produce near 
duplicates of the training data.

A common technique to assess memorization in deep generative models is to take 
samples from the model and compare these to their nearest neighbors in the 
training set. There are several problems with this approach.  First, it has 
been well established that when using the Euclidean metric this test can be 
easily fooled by taking an image from the training set and shifting it by a 
few pixels \cite{theis2016note}.  For this reason, nearest neighbors in the 
feature space of a secondary model are sometimes used, as well as cropping 
and/or downsampling before identifying nearest neighbors 
(e.g.,~\cite{karras2018progressive,brock2018large,vahdat2020nvae}). Second, 
while there may not be any neighbors in the training set for a small selection 
of samples from the model, this does not demonstrate that there are \emph{no} 
observations that are highly memorized. Indeed, in several recent publications 
on deep generative models it is possible to identify observations highly 
similar to the training set in the illustrations of generated samples (see 
Supplement~\ref{app:copies}).

Memorization in generative models is not always surprising. When the training 
data set contains a number of highly similar observations, such as duplicates, 
then it would be expected that these receive an increased weight in the model 
and are more likely to be generated. The fact that commonly-used data sets 
contain numerous (near) duplicates \cite{barz2020do} therefore provides one 
reason for memorization of training observations. While important, 
memorization due to duplicates is not the focus of this work.  Instead, we are 
concerned with memorization that arises as \emph{an increased probability of 
	generating a sample that closely resembles the training data in 
	regions of the input space where the algorithm has not seen sufficient 
	observations to enable generalization}. For example, we may expect 
that highly memorized observations are either in some way atypical or are 
essential for properly modeling a particular region of the data manifold.

\begin{figure}[tb]
	\centering
	\captionsetup[subfigure]{justification=centering}%
	\begin{subfigure}[b]{.40\textwidth}
		\centering
		\includegraphics[width=\textwidth]{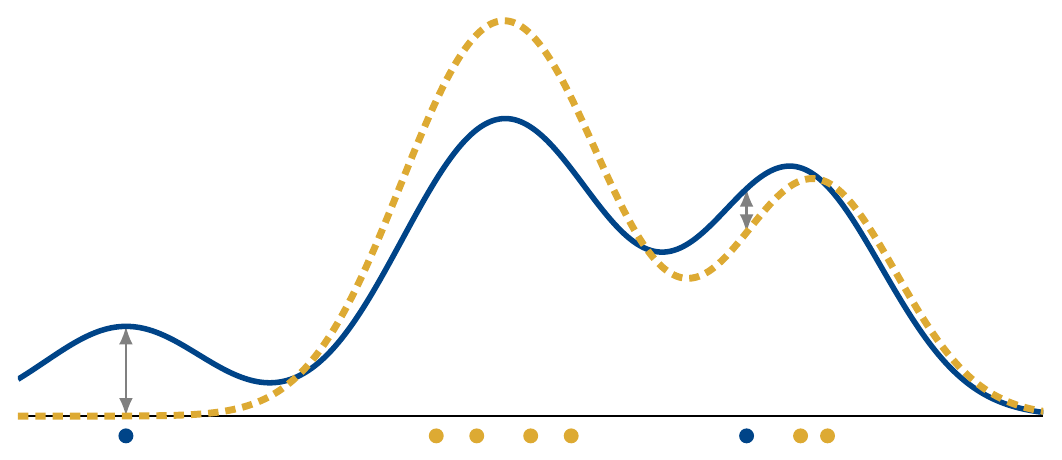}
		\caption{Memorization \label{fig:sketch_mem}}
	\end{subfigure}
	\qquad
	\begin{subfigure}[b]{.40\textwidth}
		\centering
		\includegraphics[width=\textwidth]{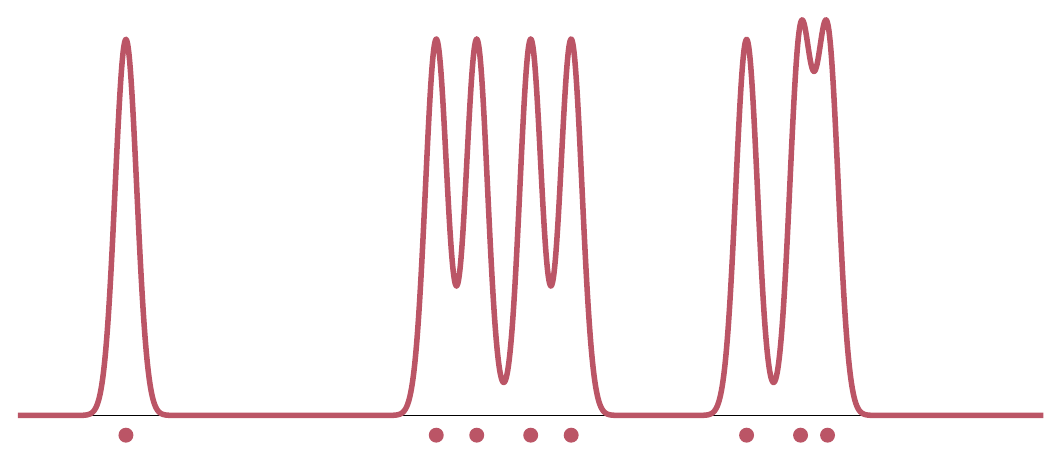}
		\caption{Overfitting \label{fig:sketch_over}}
	\end{subfigure}
	\caption{Memorization and overfitting. In (a) the solid 
		\textcolor{MyBlue}{blue} curve reflects the probability 
		density when all observations are included, whereas the dashed   
		\textcolor{MyYellow}{yellow} curve is the density when only 
		yellow observations are included. The local change in density 
		that occurs when an observation is removed from the training 
		data indicates the extent to which the model memorized the 
		observation. The density typically associated with overfitting 
		due to overtraining is shown by the solid 
		\textcolor{MyRed}{red} curve in (b).\label{fig:illustration}}%
\end{figure}

Figure~\ref{fig:sketch_mem} illustrates this kind of local memorization in 
probabilistic generative models. We focus on explicit density models as these 
are more amenable to a direct analysis of the learned probability distribution 
(as opposed to \emph{implicit} density models such as GANs). The figure shows 
that in certain regions of the input space the learned probability density can 
be entirely supported by a single, potentially outlying, observation. When 
sampling from the model in these parts of the space it is thus highly likely 
that a sample similar to an input observation will be generated. The figure 
also illustrates that in regions of the input space that are densely supported 
by closely-related observations, sampling will yield observations that 
resemble the input data.  The change in the probability density of an 
observation that occurs when it is removed from the data forms the basis of 
the memorization score we propose in Section~\ref{sec:memorization_score}.  
This form of memorization should be contrasted with what is commonly 
associated with memorization due to overfitting (illustrated in 
Figure~\ref{fig:sketch_over}). Overfitting is a \emph{global} property of a 
model that typically occurs when it is trained for too long or with too high a 
learning rate (i.e.,~\emph{overtraining}), so that a gap develops between the 
training and test performance. Thus, we emphasize that in generative models 
memorization and generalization can occur simultaneously at distinct regions 
of the input space, and that memorization is not necessarily caused by 
overtraining.

To understand memorization further, consider the simple case of fitting a 
multivariate normal distribution. In this scenario, the presence or absence of 
a particular observation in the data set will have a small effect on the 
learned model unless the observation is an outlier. By contrast, a kernel 
density estimate (KDE) \cite{rosenblatt1956remarks,parzen1962estimation} of 
the probability density may be more sensitive to the presence or absence of a 
particular observation.  To see why this is the case, consider that in 
sparsely-populated regions of the input space the KDE can be supported by a 
relatively small number of observations. Although deep generative models 
typically operate in much higher dimensional spaces than the aforementioned 
methods, the same problem can arise when generalizing to regions of the space 
that are weakly supported by the available data. Because these models are 
optimized globally, the model has to place some probability mass in these 
regions. As we will demonstrate below, it is not necessarily the case that the 
model places low probability on such  observations, resulting in observations 
that are both highly memorized and not significantly less likely under the 
model than other observations.

In this work, we extend a recently proposed measure of memorization for 
supervised learning \cite{feldman2019does,feldman2020what} to probabilistic 
generative models and introduce a practical estimator of this memorization 
score.  We subsequently investigate memorization experimentally, where we 
focus on the variational autoencoder. In our experiments we demonstrate that 
highly memorized observations are not necessarily outliers, that memorization 
can occur early during the training process, and show the connection between 
nearest neighbor tests for memorization and the proposed memorization score.  
Finally, we discuss approaches that can limit memorization in practice.

\section{Related Work}%
\label{sec:related_work}

Here we review work on memorization in deep learning, memorization as it 
relates to membership inference, the evaluation of generative models, as well 
as influence functions and stability.

\paragraph{Memorization in deep learning.} The observation that deep learning 
models can learn from patterns of random data has been a catalyst for recent 
efforts to understand memorization in supervised learning 
\cite{zhang2017understanding,arpit2017closer,stephenson2021on}. A number of 
approaches have been proposed to test for memorization in specific 
applications. In \cite{carlini2019secret} memorization in language models is 
evaluated using a ``canary'' string (e.g.,~if ``my social security number is 
$x$'' is in the training data, how often does the model complete the prompt 
``my social security number is'' using $x$ instead of a comparable $y \neq x$ 
that is not in the training set).  Unfortunately, this approach does not 
translate easily to other contexts, such as images. Moreover, language models 
often contain explicit memory cells such as LSTMs \cite{hochreiter1997long} 
that can facilitate memorization, which are absent in most generative models.

A memorization score for supervised learning was proposed in 
\cite{feldman2019does}, which forms the inspiration for our formulation in 
Section~\ref{sec:memorization_score}. A related ``consistency score'' for 
supervised learning was proposed in \cite{jiang2021characterizing}.  We argue, 
however, that memorization in supervised learning differs fundamentally from 
that in generative models, as the label prediction task affects the training 
dynamics, and label noise is known to induce memorization in supervised 
learning \cite{liu2020early,xia2021robust}. Building on earlier work by 
\cite{lopez2016revisiting,xu2018empirical}, a hypothesis test is proposed in 
\cite{meehan2020non} that is based on the premise that memorization has 
occurred when samples from the trained model are ``closer'' to the training 
data than observations from the test set. While this is a useful test for 
aggregate memorization behavior in (a region of) the input space, our proposed 
score function allows us to quantify the memorization of a single observation.

\paragraph{Membership inference.} A topic closely related to memorization is 
the problem of membership inference. Here, the goal is to recover whether a 
particular observation was part of the unknown training data set, either using 
knowledge of the model, access to the model, or in a black-box setting.  
Membership inference is particularly important when models are deployed 
\cite{fredrikson2015model}, as potentially private data could be exposed. In 
the supervised learning setting, \cite{shokri2017membership} propose to use an 
attack model that learns to classify whether a given sample was in the 
training set. Later work \cite{hitaj2017deep,hayes2019logan} focused on 
generative models and proposed to train a GAN on samples from the target 
model.  The associated discriminator is subsequently used to classify 
membership of the unknown training set. A related approach to recovering 
training images is described in \cite{webster2019detecting}, using an 
optimization algorithm that identifies for every observation the closest 
sample that can be generated by the network.  However this requires solving a 
highly non-convex problem, which isn't guaranteed to find the optimal 
solution.

\paragraph{Evaluating generative models.} Memorization is a known issue when  
evaluating generative models, in particular for GANs  
\cite{arora2018do,xu2018empirical}. Several approaches are discussed in 
\cite{theis2016note}, with a focus on the pitfalls of relying on 
log-likelihood, sample quality, and nearest neighbors. Using the 
log-likelihood can be particularly problematic as it has been shown that 
models can assign higher likelihood to observations outside the input domain 
\cite{nalisnick2019deep}.  Nowadays, generative models are frequently 
evaluated by the quality of their samples as evaluated by other models, as is 
done in the Inception Score (IS) \cite{salimans2016improved} and Fr\'echet 
Inception Distance (FID) \cite{heusel2017gans}. Since these metrics have no 
concept of where the samples originate, the pathological case where a model 
memorizes the entire training data set will yield a near-perfect score.  
Motivated by this observation, \cite{gulrajani2018towards} propose to use 
neural network divergences to measure sample diversity and quality 
simultaneously, but this requires training a separate evaluation model and 
there is no guarantee that \emph{local} memorization will be detected.

\paragraph{Influence \& Stability.} The problem of memorization is also 
related to the concept of \emph{influence functions} in statistics 
\cite{hampel1974influence,cook1980characterizations}. Influence functions can 
be used to measure the effect of upweighting or perturbing an observation and 
have recently been considered as a diagnostic tool for deep learning 
\cite{koh2017understanding,terashita2021influence}.  However, it has also been 
demonstrated that influence function estimates in deep learning models can be 
fragile \cite{basu2021influence}. Below, we therefore focus on a relatively 
simple estimator to gain a reliable understanding of memorization in 
probabilistic deep generative models. Concurrent work in 
\cite{kong2021understanding} focuses on influence functions for variational 
autoencoders based on the ``TracIn'' approximation of 
\cite{pruthi2020estimating} computed over the training run. An important 
difference is that while the method of \cite{kong2021understanding} is 
computed for one particular model, our memorization score applies to a 
particular model \emph{architecture} by averaging over multiple $K$-fold 
cross-validation fits (see Section~\ref{sec:memorization_score}). Related to 
influence functions is the concept of stability in learning theory 
\cite{bousquet2002stability}. In particular, the \emph{point-wise hypothesis 
	stability} is an upper bound on the expected absolute change in the 
loss function when an observation is removed from the training set (where the 
expectation is over all training sets of a given size). We instead focus on 
the change in the density of a probabilistic model when trained on a specific 
data set.

\section{Memorization Score}%
\label{sec:memorization_score}

We present a principled formulation of a memorization score for probabilistic 
generative models, inspired by the one proposed recently in 
\cite{feldman2019does,feldman2020what} for supervised learning. Let 
$\mathcal{A}$ denote a randomized learning algorithm, and let $a$ be an 
instance of the algorithm (i.e.,~a trained model). Here, $\mathcal{A}$ 
captures a complete description of the algorithm, including the chosen 
hyperparameters, training epochs, and optimization method. The randomness in 
$\mathcal{A}$ arises from the particular initial conditions, the selection of 
mini batches during training, as well as other factors. Denote the training 
data set by $\mathcal{D} = \{\bft{x}_i\}_{i=1}^n$ with observations from 
$\mathcal{X} \subseteq \mathbb{R}^D$.  Let $[n] = \{1, \ldots, n\}$ and write 
$\mathcal{D}_{\mathcal{I}} = \{\bft{x}_i : \bft{x}_i \in \mathcal{D}, i \in 
\mathcal{I}\}$ for the subset of observations in the training data indexed by 
the set $\mathcal{I} \subseteq [n]$.  The posterior probability assigned to an 
observation $\bft{x} \in \mathcal{X}$ by a model $a$ when trained on a data 
set $\mathcal{D}$ is written as $p(\bft{x} \given \mathcal{D}, a)$.

We are interested in the posterior probability of an observation assigned by 
the algorithm $\mathcal{A}$, not merely by an instantiation of the algorithm.  
Therefore we introduce the probability
$P_{\mathcal{A}}(\bft{x} \given \mathcal{D})$ and its sampling estimate as
\begin{equation}
	\label{eq:logpAx}
	P_{\mathcal{A}}(\bft{x} \given \mathcal{D}) = \int p(\bft{x} \given 
	\mathcal{D}, a) p(a) \myd a \approx \frac{1}{T} \sum_{t = 1}^T 
	p(\bft{x} \given \mathcal{D}, a_t),
\end{equation}
for some number of repetitions $T$. We see that $P_{\mathcal{A}}(\bft{x} 
\given \mathcal{D})$ is the expectation of $p(\bft{x} \given \mathcal{D}, a)$ 
over instances of the randomized algorithm $\mathcal{A}$.

To facilitate meaningful interpretation of the memorization score we use the 
difference in log probabilities, in contrast to 
\cite{feldman2019does,feldman2020what}. Thus we define the leave-one-out (LOO) 
memorization score as
\begin{equation}
	\label{eq:mem_exact}
	M^{\text{LOO}}(\mathcal{A}, \mathcal{D}, i) = \log 
	P_{\mathcal{A}}(\bft{x}_i \given \mathcal{D}) - \log 
	P_{\mathcal{A}}(\bft{x}_i \given \mathcal{D}_{[n] \setminus \{i\}}).
\end{equation}
This memorization score measures how much more likely an observation is when 
it is included in the training set compared to when it is not. For example, if 
$M^{\text{LOO}}(\mathcal{A}, \mathcal{D}, i) = 10$, then 
$P_{\mathcal{A}}(\bft{x}_i \given \mathcal{D}) = \exp(10) \cdot 
P_{\mathcal{A}}(\bft{x}_i \given \mathcal{D}_{[n] \setminus \{i\}})$.  
Moreover, when $M^{\text{LOO}}(\mathcal{A}, \mathcal{D}, i) = 0$ removing the 
observation from the training data has no effect at $\bft{x}_i$, and when 
$M^{\text{LOO}}(\mathcal{A}, \mathcal{D}, i) < 0$ the observation is more 
likely under the model when it is removed from the training data. We will 
abbreviate the LOO memorization score as $M^{\text{LOO}}_i := 
M^{\text{LOO}}(\mathcal{A}, \mathcal{D}, i)$ when the arguments are clear from 
context.

\paragraph{Estimation.} The memorization score in (\ref{eq:mem_exact}) is a 
leave-one-out estimator that requires fitting the learning algorithm 
$\mathcal{A}$ multiple times for each observation as it is left out of the 
training data set. As this is computationally infeasible in general, we 
introduce a practical estimator that simplifies the one proposed in 
(\ref{eq:mem_exact}). Instead of using a leave-one-out method or random 
sampling, we use a $K$-fold approach as is done in cross-validation.  Let 
$\mathcal{I}_k$ denote randomly sampled disjoint subsets of the indices $[n] = 
\{1,\ldots,n\}$ of size $n / K$, such that $\cup_{k=1}^K \mathcal{I}_k = [n]$.  
We then train the model on each of the training sets $\mathcal{D}_{[n] 
	\setminus \mathcal{I}_k}$ and compute the log probability for all 
observations in the training set and the holdout set 
$\mathcal{D}_{\mathcal{I}_k}$.

Since there is randomness in the algorithm $\mathcal{A}$ and in the chosen 
folds $\mathcal{I}_k$, we repeat the cross-validation procedure $L$ times and 
average the results. Writing $\mathcal{I}_{\ell, k}$ for the $k$-th holdout 
index set in run $\ell$ and abbreviating the respective training set as 
$\mathcal{D}_{\ell, k} = \mathcal{D}_{[n] \setminus \mathcal{I}_{\ell,k}}$,
the memorization score becomes
\begin{align}
	\label{eq:mem_approx}
	M^{\text{K-fold}}_{i} &= \log \left[ \frac{1}{L(K-1)} \sum_{\ell=1}^L 
		\sum_{k=1}^K \bbone_{i \notin \mathcal{I}_{\ell, k}} 
		p(\bft{x}_i \given \mathcal{D}_{\ell,k}, a_{\ell, k}) \right] 
	\\
	&\quad - \log \left[ \frac{1}{L} \sum_{\ell=1}^L \sum_{k=1}^K 
		\bbone_{i \in \mathcal{I}_{\ell, k}} p(\bft{x}_i \given 
		\mathcal{D}_{\ell,k}, a_{\ell, k}) \right], \nonumber
\end{align}
where $\bbone_{v}$ is the indicator function that equals $1$ if $v$ is true 
and $0$ otherwise. Each of the $K-1$ folds where observation $i$ is in the 
training set contributes to the first term in (\ref{eq:mem_exact}), and when 
observation $i$ is in the holdout set it then contributes to the second term.
This approach is summarized in Algorithm~\ref{alg:mem}, where log 
probabilities are used for numerical accuracy. In practice, the number of 
repetitions $L$ and folds $K$ will be dominated by the available computational 
resources. 

\begin{algorithm}[tb]
	\algrenewcommand\algorithmicrequire{\textbf{Input:}}
	\algrenewcommand\algorithmicensure{\textbf{Output:}}
	\caption{\small Computing the Cross-Validated Memorization Score 
		\label{alg:mem}}
	\begin{algorithmic}[1]
		\small
		\Require Algorithm $\mathcal{A}$, data set $\mathcal{D}$, 
		repetitions $L$, folds $K$
		\Ensure $M^{\text{K-fold}}_{i}$, $\forall i$
		\For {$\ell = 1, \ldots, L$}
			\State $\mathcal{G}_{\ell} \gets$ Random partition of 
			$[n]$ into $K$ disjoint subsets
			\For {$\mathcal{I}_{\ell,k} \in \mathcal{G}_{\ell}$ 
				with $k = 1,\ldots, K$}
				\State $a_{\ell, k} \gets$ Train $\mathcal{A}$  
				on $\mathcal{D}_{[n] \setminus 
					\mathcal{I}_{\ell, k}}$
				\State $\pi_{\ell, k, i} \gets $ Compute $\log 
				p(\bft{x}_i \given \mathcal{D}_{[n] \setminus 
					\mathcal{I}_{\ell, k}}, a_{\ell, k})$, 
				$\forall i \in [n]$
			\EndFor
		\EndFor
		\Comment \textcolor{gray}{\scriptsize 
			\Call{LogMeanExp}{$\{u_i\}_{i=1}^{n}$} = $-\log n + 
			\Call{LogSumExp}{\{ u_i \}_{i=1}^{n} }$}
		\vskip.2\baselineskip
		\State $U_i \gets$ \Call{LogMeanExp}{$\{ \pi_{\ell, k, i} : 
			\ell \in [L], k \in [K], i \notin \mathcal{I}_{\ell, 
				k} \}$}, $\forall i$
		\State $V_i \gets$ \Call{LogMeanExp}{$\{ \pi_{\ell, k, i} : 
			\ell \in [L], k \in [K], i \in \mathcal{I}_{\ell, k} 
			\}$}, $\forall i$
		\State $M^{\text{K-fold}}_i \gets U_i - V_i$, $\forall i$
	\end{algorithmic}
\end{algorithm}

\paragraph{When is memorization significant?} A natural question is what 
values of the memorization score are significant and of potential concern. The 
memorization scores can be directly compared between different algorithm 
settings on the same data set, for instance to understand whether changes in 
hyperparameters or model architectures increase or decrease memorization.  
Statistical measures such as the mean, median, and skewness of the 
memorization score or the location of, say, the 95th percentile, can be 
informative when quantifying memorization of a particular model on a 
particular data set, but can not necessarily be compared between data sets.  
In practice, we also find that the distribution of the memorization score can 
differ between modes in the data set, such as distinct object classes. This 
can be understood by considering that the variability of observations of 
distinct classes likely differs, which affects the likelihood of the objects 
under the model, and in turn the memorization score. We will return to this 
question in Section~\ref{sec:discussion}.

\section{Experiments}
\label{sec:experiments}

We next describe several experiments that advance our understanding of 
memorization in probabilistic deep generative models, with a focus on the 
variational autoencoder setting. Additional results are available in 
Supplement~\ref{app:additional_results}. Code to reproduce our experiments can 
be found in an online repository.\footnote{See: 
	\url{https://github.com/alan-turing-institute/memorization}.}

\subsection{Background}

We employ the variational autoencoder (VAE) 
\cite{kingma2014auto,rezende2014stochastic} as the probabilistic generative 
model in our experiments, although it is important to emphasize that the 
memorization score introduced above is equally applicable to methods such as 
normalizing flows, diffusion networks, and other generative models that learn 
a probability density over the input space. The VAE is a latent-variable 
model, where we model the joint distribution $p_{\theta}(\bft{x}, \bft{z})$ of 
an observation $\bft{x} \in \mathcal{X} \subseteq \mathbb{R}^D$ and a latent 
variable $\bft{z} \in \mathcal{Z} \subseteq \mathbb{R}^{d}$. The joint 
distribution can be factorized as $p_{\theta}(\bft{x}, \bft{z}) = 
p_{\theta}(\bft{x} \given \bft{z}) p(\bft{z})$, and in the VAE the prior 
distribution $p(\bft{z})$ is typically assumed to be a standard multivariate 
Gaussian. The posterior distribution $p_{\theta}(\bft{z} \given \bft{x})$ is 
generally intractable, so it is approximated using an inference model, or 
\emph{encoder}, $q_{\phi}(\bft{z} \given \bft{x})$. Analogously, the model 
$p_{\theta}(\bft{x} \given \bft{z})$ is often referred to as the 
\emph{decoder}. The VAE is trained by maximizing the lower bound on the 
evidence (ELBO), see (\ref{eq:elbo}), since
\begin{align}
	\log p_{\theta}(\bft{x}) &\geq \mathbb{E}_{q_{\phi}(\bft{z} \given 
		\bft{x})} \left[ \log p_{\theta}(\bft{x}, \bft{z}) - \log 
		q_{\phi}(\bft{z} \given \bft{x}) \right] \\
	&= - D_{\text{KL}}(q_{\phi}(\bft{z} \given \bft{x}) \,\|\, p(\bft{z})) 
	+ \mathbb{E}_{q_{\phi}(\bft{z} \given \bft{x})}\left[ \log 
		p_{\theta}(\bft{x} \given \bft{z}) \right] \label{eq:elbo},
\end{align}
with $D_{\text{KL}}(\cdot \,\|\, \cdot)$ the Kullback-Leibler (KL) divergence 
\cite{kullback1951information}. By choosing a simple distribution for the 
encoder $q_{\phi}(\bft{z} \given \bft{x})$, such as a multivariate Gaussian, 
the KL divergence has a closed-form expression, resulting in an efficient 
training algorithm.

We use importance sampling on the decoder \cite{burda2016importance} to 
approximate $\log p_{\theta}(\bft{x}_i)$ for the computation of the 
memorization score, and focus on the MNIST \cite{lecun1998mnist}, CIFAR-10 
\cite{krizhevsky2009learning}, and CelebA \cite{liu2015faceattributes} data 
sets. We use a fully connected encoder and decoder for MNIST and employ 
convolutional architectures for CIFAR-10 and CelebA. For the optimization we 
use Adam \cite{kingma2015adam} and we implement all models in PyTorch 
\cite{paszke2019pytorch}. The memorization score is estimated using $L = 10$ 
repetitions and $K = 10$ folds. Additional details of the experimental setup 
and model architectures can be found in Supplement~\ref{app:details}.

\subsection{Results}
\label{sub:results}

\begin{figure}[tb]
	\centering
	\captionsetup[subfigure]{justification=centering}%
	\begin{subfigure}[b]{.30\textwidth}
		\includegraphics[width=\textwidth]{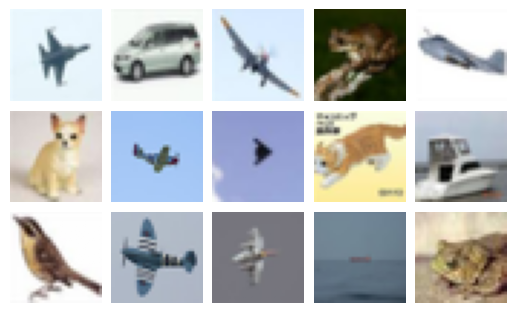}
		\caption{Low memorization \label{fig:cifar10_bottom}}
	\end{subfigure}
	\quad
	\begin{subfigure}[b]{.30\textwidth}
		\includegraphics[width=\textwidth]{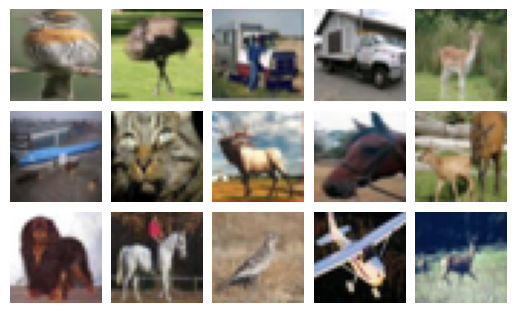}
		\caption{Median memorization \label{fig:cifar10_middle}}
	\end{subfigure}
	\quad
	\begin{subfigure}[b]{.30\textwidth}
		\includegraphics[width=\textwidth]{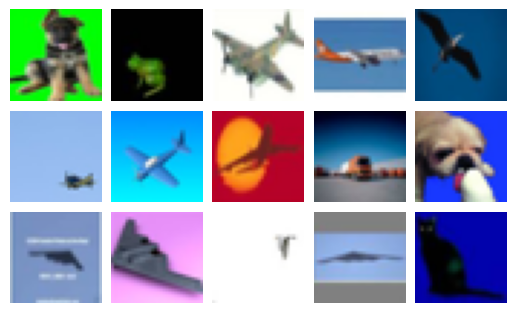}
		\caption{High memorization \label{fig:cifar10_top}}
	\end{subfigure}
	\caption{Observations with low, median, and high memorization scores 
		in the CIFAR-10 data set, when learning the distribution with 
		a convolutional VAE. Memorization scores range from about 
		$-180$ in the top left of figure (a) to about $900$ in the 
		bottom right of figure (c), with a median of $97$.  
		\label{fig:cifar10_mem}}
\end{figure}

We first explore memorization qualitatively. Figure~\ref{fig:cifar10_mem} 
shows examples of observations with low, median, and high memorization scores 
in the VAE model trained on CIFAR-10. While some of the highly-memorized 
observations may stand out as odd to a human observer, others appear not 
unlike those that receive a low memorization score. This shows that the kind 
of observations that are highly memorized in a particular model may be 
counterintuitive, and are not necessarily visually anomalous.

\begin{figure}[tb]
	\centering
	\captionsetup[subfigure]{justification=centering}%
	\begin{subfigure}[b]{0.51\textwidth}
		\includegraphics[height=48mm]{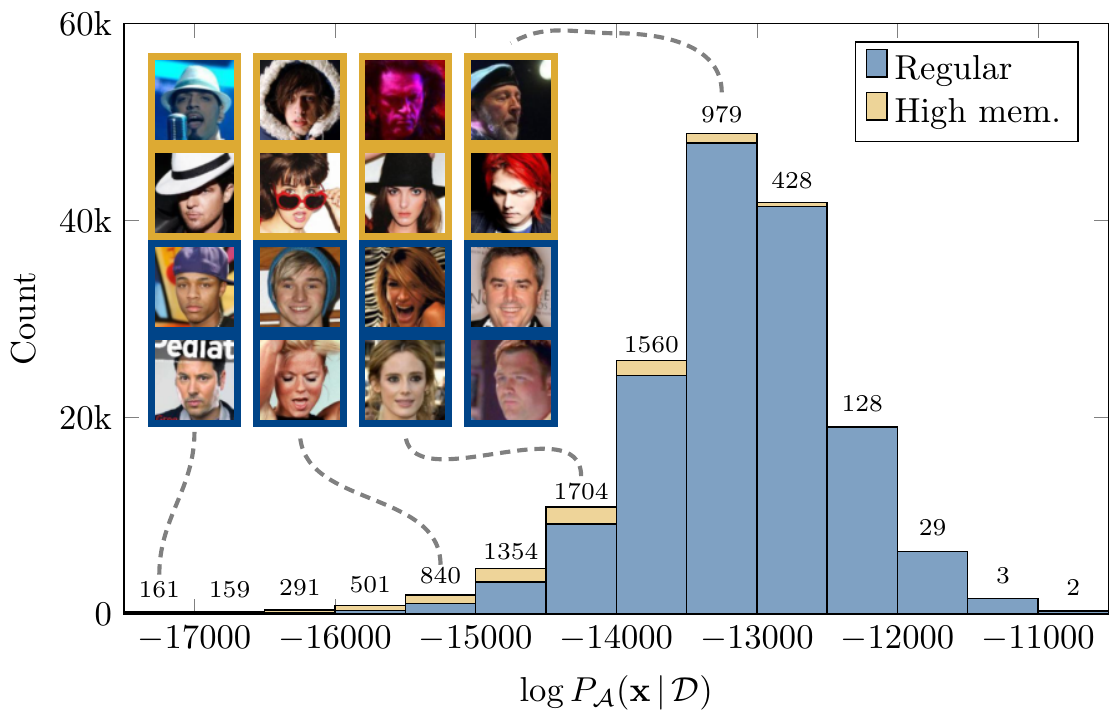}
		\caption{Counts \label{fig:celeba_logpx_bins}}
	\end{subfigure}
	\qquad
	\begin{subfigure}[b]{0.43\textwidth}
		\includegraphics[height=48mm]{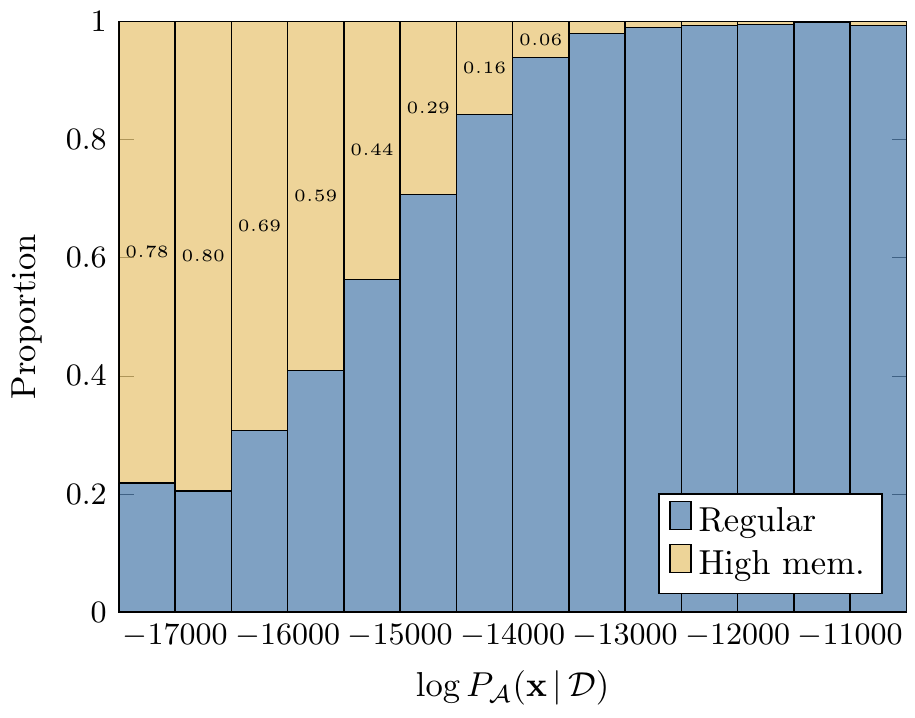}
		\caption{Proportions \label{fig:celeba_logpx_prop}}
	\end{subfigure}
	\caption{In (a) we show a histogram of the number of highly memorized 
		(\textcolor{MyYellow}{yellow}) and regular 
		(\textcolor{MyBlue}{blue}) observations for bins of the log 
		probability under a VAE model trained on the CelebA data set 
		(where $n = 162,770$).  The numbers above the bars correspond 
		to the number of highly memorized observations in each bin.  
		Randomly selected training observations from several bins are 
		shown as well, with dashed lines illustrating the bin where 
		the images in a particular column can be found. Images with a 
		yellow frame are highly memorized whereas those with a blue 
		frame have low memorization scores. Figure (b) shows the 
		\emph{proportion} of highly memorized and regular observations 
		for each bin.
		\label{fig:celeba_mem}}
\end{figure}

If highly memorized observations are always given a low probability when they 
are included in the training data, then it would be straightforward to dismiss 
them as outliers that the model recognizes as such. However, we find that this 
is not universally the case for highly memorized observations, and a sizable 
proportion of them are likely \emph{only} when they are included in the 
training data. If we consider observations with the 5\% highest memorization 
scores to be ``highly memorized'', then we can check how many of these 
observations are considered likely by the model when they are included in the 
training data.  Figure~\ref{fig:celeba_logpx_bins} shows the number of highly 
memorized and ``regular'' observations for bins of the log probability under 
the VAE model for CelebA, as well as example observations from both groups for 
different bins. Moreover, Figure~\ref{fig:celeba_logpx_prop} shows the 
proportion of highly memorized observations in each of the bins of the log 
probability under the model.  While the latter figure shows that observations 
with low probability are \emph{more likely} to be memorized, the former shows 
that a considerable proportion of highly memorized observations are \emph{as 
	likely as regular observations} when they are included in the training 
set.  Indeed, more than half the highly memorized observations fall within the 
central 90\% of log probability values (i.e.,~with $\log 
P_{\mathcal{A}}(\bft{x} \given \mathcal{D}) \in [-14500, -12000]$).

The memorization score can be a useful diagnostic tool to evaluate the effect 
of different hyperparameter settings and model architectures. For example, in 
Figure~\ref{fig:mnist_mem_lrs_mem} we illustrate the distribution of the 
memorization score for a VAE trained on MNIST with two different learning 
rates, and we show the train and test set losses during training in 
Figure~\ref{fig:mnist_mem_lrs_loss}. With a learning rate of $\eta = 10^{-3}$ 
(blue curves), a clear generalization gap can be seen in the loss curves, 
indicating the start of overtraining (note the test loss has not yet started 
to \emph{increase}). This generalization gap disappears when training with the 
smaller learning rate of $\eta = 10^{-4}$ (yellow curves).  The absence of a 
generalization gap is sometimes used as evidence for the absence of 
overfitting and memorization \cite{wu2016quantitative}, but the distribution 
of the memorization scores for $\eta = 10^{-4}$ in 
Figure~\ref{fig:mnist_mem_lrs_mem} shows that this is insufficient.  While the 
memorization scores are \emph{reduced} by lowering the learning rate, 
relatively high memorization values can still occur. In fact, the largest 
memorization score for $\eta = 10^{-4}$ is about 80, and represents a shift 
from a far outlier when the observation is absent from the training data to a 
central inlier when it is present.\footnote{For this particular observation, 
	$\log P_{\mathcal{A}}(\bft{x}_i \given \mathcal{D}_{[n] \setminus 
		\{i\}}) \approx -178$ when the observation is excluded from 
	the training data, and $\log P_{\mathcal{A}}(\bft{x}_i \given 
	\mathcal{D}) \approx -97$ when it is included, and the latter value is 
	approximately equal to the average log probability of the other 
	observations with the same digit.}

\subsection{Memorization during training}

\begin{figure}[tb]
	\centering
	\def\FigSizeMNIST{0.329\textwidth}
	\captionsetup[subfigure]{justification=centering}%
	\begin{subfigure}[b]{\FigSizeMNIST}
		\includegraphics[height=50mm]{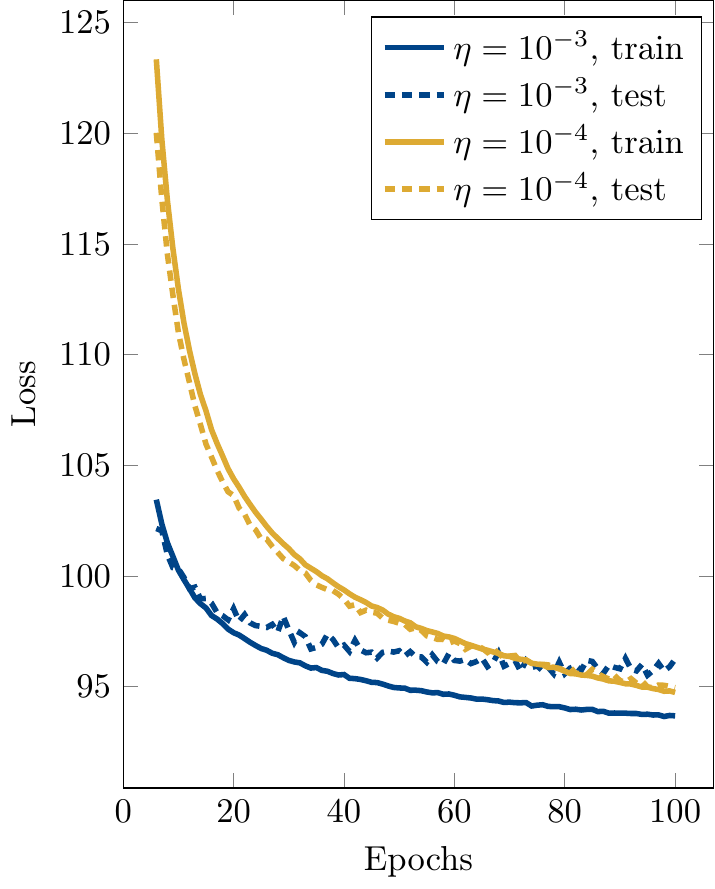}
		\caption{Loss curves \label{fig:mnist_mem_lrs_loss}}
	\end{subfigure}
	\begin{subfigure}[b]{\FigSizeMNIST}
		\includegraphics[height=50mm]{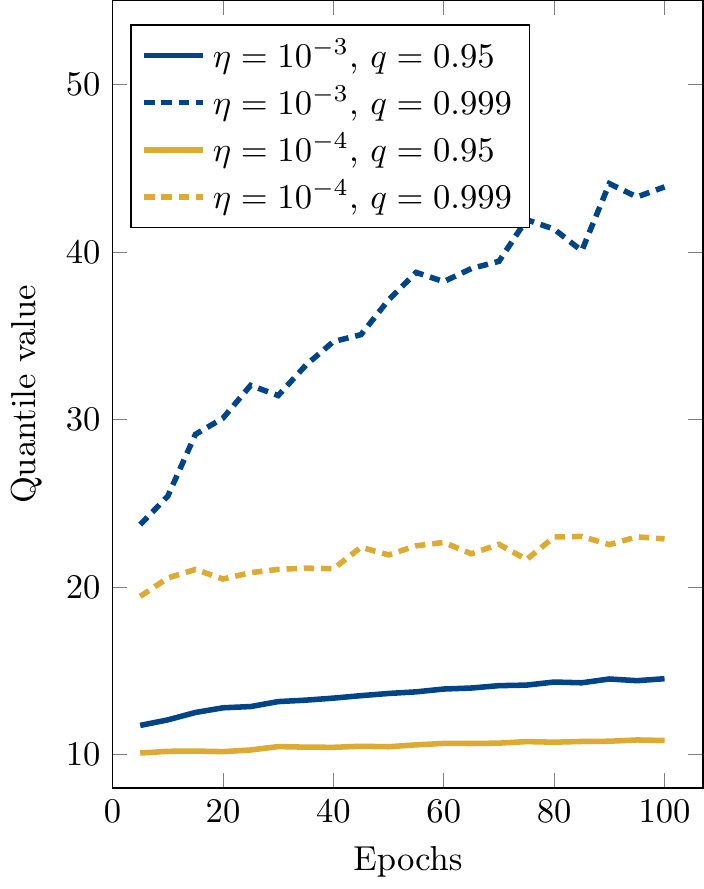}
		\caption{Memorization quantiles 
			\label{fig:mnist_mem_lrs_quantiles}}
	\end{subfigure}
	\begin{subfigure}[b]{\FigSizeMNIST}
		\includegraphics[height=50mm]{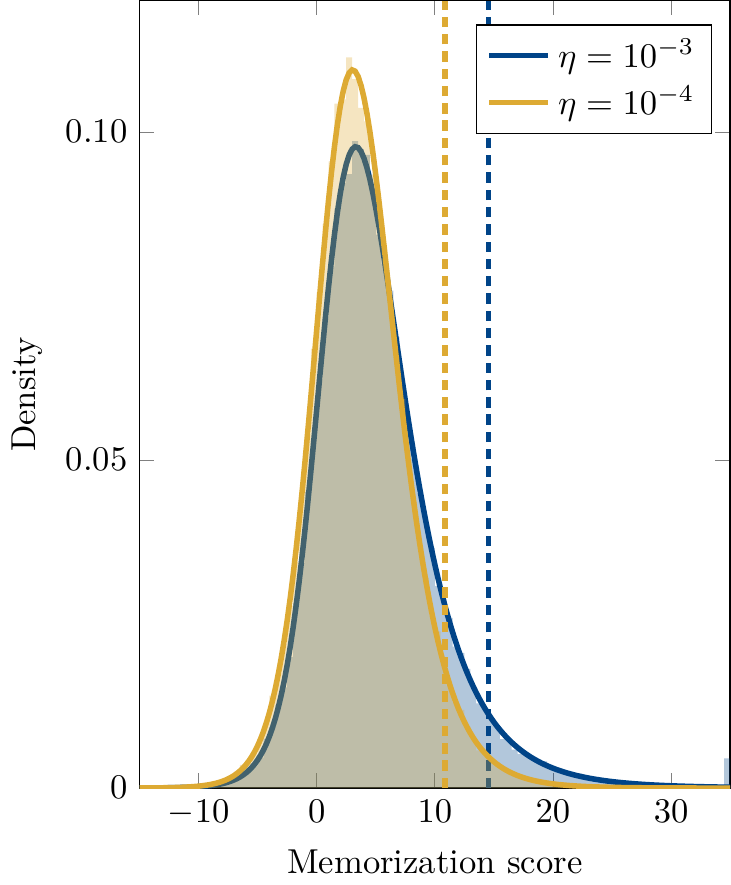}
		\caption{Memorization score \label{fig:mnist_mem_lrs_mem}}
	\end{subfigure}
	\caption{Loss curves, memorization score quantiles, and memorization 
		score distributions for a VAE trained on MNIST with different 
		learning rates, $\eta = 10^{-3}$ (\textcolor{MyBlue}{blue}) 
		and $\eta = 10^{-4}$ (\textcolor{MyYellow}{yellow}). In (c) 
		the vertical lines mark the 95th percentile of the 
		memorization scores, and the axes are cropped for clarity (the 
		maximum memorization score for $\eta = 10^{-3}$ is about 210 
		and for $\eta = 10^{-4}$ it is about 80).  
		\label{fig:mnist_mem_lrs}}
\end{figure}

To continue on the relation between memorization and overtraining, we look at 
how the memorization score evolves during training. In 
Figure~\ref{fig:mnist_mem_lrs_quantiles} we show the 0.95 and 0.999 quantiles 
of the memorization score for the VAE trained on MNIST using two different 
learning rates. The quantiles are chosen such that they show the memorization 
score for the highly memorized observations. For both learning rates we see 
that the memorization score quantiles increase during training, as can be 
expected.  However we also see that for the larger learning rate of $\eta = 
10^{-3}$ the memorization score quantiles already take on large values before 
the generalization gap in Figure~\ref{fig:mnist_mem_lrs_loss} appears.  This 
is additional evidence that determining memorization by the generalization gap 
is insufficient, and implies that early stopping would not fully alleviate 
memorization. Moreover, we see that the rate of increase for the peak 
memorization quantiles slows down with more training, which suggests that the 
memorization score stabilizes and does not keep increasing with the training 
epochs. This is reminiscent of \cite{carlini2019secret}, who demonstrated that 
their metric for memorization in language models peaks when the test loss 
starts to increase. The difference is that here memorization appears to 
stabilize even before this happens.

\subsection{Nearest Neighbors}
\label{sub:nearest_neighbors}

As discussed in the introduction, nearest neighbor illustrations are commonly 
used to argue that no memorization is present in the model. Moreover, 
hypothesis tests and evaluation metrics have been proposed that measure 
memorization using distances between observations and model samples
\cite{xu2018empirical,meehan2020non}. Because of the prevalence of nearest 
neighbor tests for memorization, we next demonstrate the relationship between 
our proposed memorization score and a nearest neighbor metric.

As an example of a nearest neighbor test, we look at the relative distance of 
observations from the training set to generated samples and observations from 
the validation set. Let $\mathcal{S} \subseteq \mathcal{X}$ be a set of 
samples from the model and let $\mathcal{V} \subseteq \mathcal{X}$ be the 
validation set, with $\abs{\mathcal{S}} = \abs{\mathcal{V}}$.  Denote by $d : 
\mathcal{X} \times \mathcal{X} \rightarrow \mathbb{R}$ a distance metric, 
which we choose here to be the Euclidean distance between images in pixel 
space after downsampling by a factor of 2. For all $\bft{x}_i$ in the training 
set $\mathcal{D}$, we then compute the ratio between the closest distance to a 
member of the validation set and a member of the sample set,
\begin{equation}
	\label{eq:nn_ratio}
	\rho_i = \frac{\min_{\bft{x} \in \mathcal{V}} d(\bft{x}_i, 
		\bft{x})}{\min_{\bft{x} \in \mathcal{S}} d(\bft{x}_i, 
		\bft{x})}.
\end{equation}
If $\rho_i > 1$, then the nearest neighbor of $\bft{x}_i$ in the sample set is 
closer than the nearest neighbor in the validation set, and vice versa. Thus 
$\rho_i > 1$ suggests memorization is occurring, but as it depends on sampling 
it is expected to be very noise at an individual data point. Investigating if 
the \emph{average} ratio for a set of observations differs significantly from 
1 is an example of using hypothesis testing approaches to measure 
memorization.

\begin{figure}[tb]
	\centering
	\captionsetup[subfigure]{justification=centering}%
	\begin{subfigure}[b]{.40\textwidth}
		\includegraphics[height=40mm]{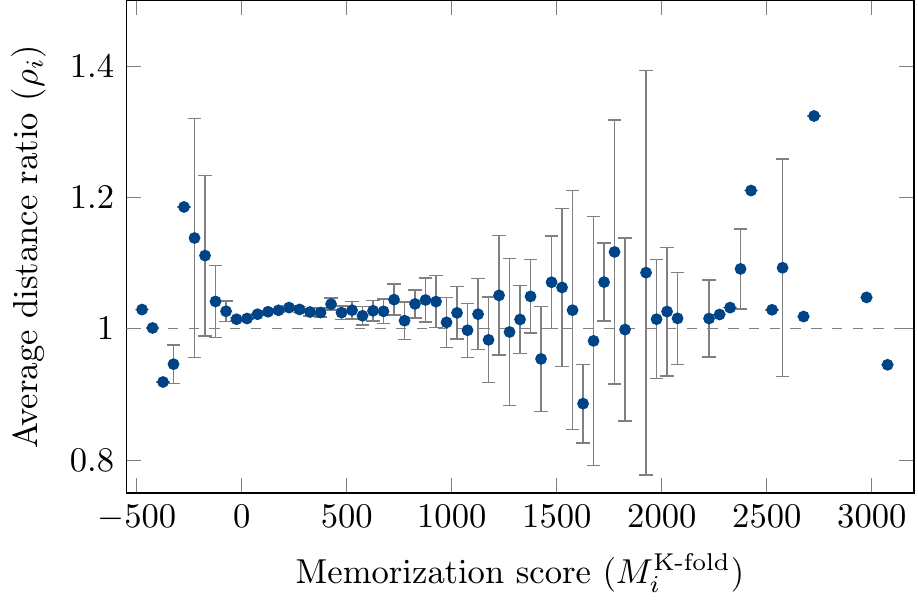}
		\caption{$M^{\text{K-fold}}_i$ vs. $\rho_i$ 
			\label{fig:nn_vs_mem_scatter}}
	\end{subfigure}
	\qquad
	\qquad
	\begin{subfigure}[b]{.40\textwidth}
		\includegraphics[height=40mm]{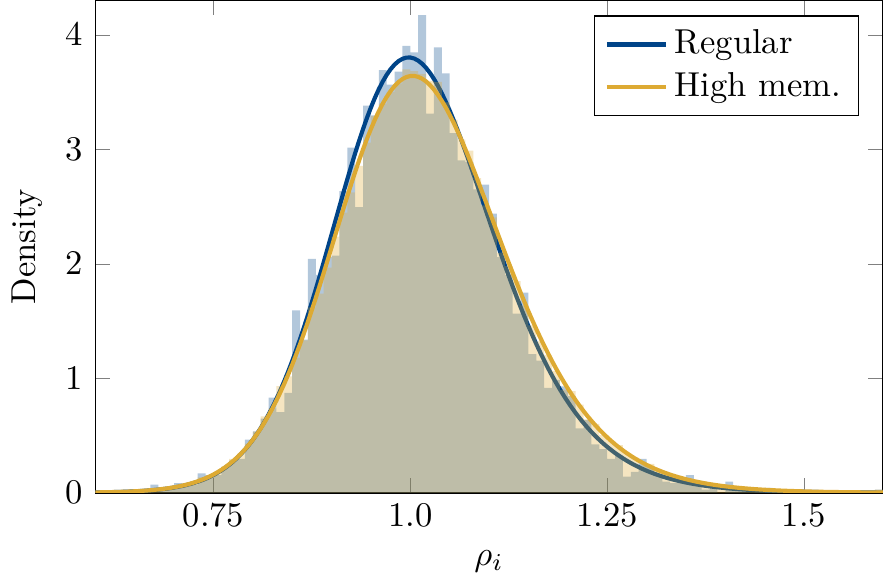}
		\caption{Distribution of $\rho_i$ \label{fig:nn_vs_mem_hist}}
	\end{subfigure}
	\caption{Nearest neighbor distance ratio in (\ref{eq:nn_ratio}) 
		compared to the memorization score for a VAE trained on 
		CelebA. Due to the large number of observations, we present 
		the average of $\rho_i$ for bins of the memorization score of 
		width $50$, and show error bars representing the confidence 
		interval of the standard error of the mean of distance ratio 
		measurements in each bin. The horizontal axis in figure~(a) 
		trims off one observation at $M^{\text{K-fold}}_i \approx 
		6500$ for clarity. Figure~(b) shows the distribution of the 
		distance ratio for observations with a high memorization score 
		(top 5\%) and the regular ones.}%
	\label{fig:nn_vs_mem}
\end{figure}

Figure~\ref{fig:nn_vs_mem} illustrates the relationship between $\rho_i$ and 
the memorization score $M^{\text{K-fold}}_i$. We see that in general there is 
no strong correlation between the two score functions, which can be explained 
by the fact that they measure different quantities. While the memorization 
score directly measures how much the model relies on the presence of 
$\bft{x}_i$ for the local probability density, nearest neighbor methods test 
how ``close'' samples from the model are to the training or validation data.  
They thus require a meaningful distance metric (which is non-trivial for 
high-dimensional data) and are subject to variability in the sample and 
validation sets.  We therefore argue that while nearest neighbor examples and 
hypothesis tests can be informative and may detect global memorization, to 
understand memorization at an instance level the proposed memorization score 
is to be preferred.

\section{Mitigation Strategies}
\label{sec:mitigation}

We describe two strategies that can be used to mitigate memorization in 
probabilistic generative models. First, the memorization score can be directly 
related to the concept of Differential Privacy (DP) 
\cite{dwork2006calibrating,dwork2014algorithmic}. Note that the memorization 
score in (\ref{eq:mem_exact}) can be rewritten as
\begin{equation}
	P_{\mathcal{A}}(\bft{x}_i \given \mathcal{D}) = \exp(M^{\text{LOO}}_i) 
	P_{\mathcal{A}}(\bft{x}_i \given \mathcal{D}_{[n] \setminus \{i\}}),
\end{equation}
and recall that a randomized algorithm $\mathcal{A}$ is 
$\varepsilon$-differentially private if for \emph{all} data sets 
$\mathcal{D}_1, \mathcal{D}_2$ that differ in only one element the following 
inequality holds
\begin{equation}
	P_{\mathcal{A}}(\mathcal{W} \given \mathcal{D}_1) \leq 
	\exp(\varepsilon) P_{\mathcal{A}}(\mathcal{W} \given 
	\mathcal{D}_2),\quad \forall\, \mathcal{W} \subseteq \mathcal{X}.
\end{equation}
Since this must hold for all subsets $\mathcal{W}$ of $\mathcal{X}$, it must 
also hold for the case where $\mathcal{W} = \{ \bft{x}_i \}$. Moreover, when 
$\bft{x}_i$ is removed from $\mathcal{D}$ it can be expected that the largest 
change in density occurs at $\bft{x}_i$. It then follows that the memorization 
score can be bounded by employing $\varepsilon$-DP estimation techniques when 
training the generative model, as this will guarantee that $M^{\text{LOO}}_i 
\leq \varepsilon, \forall i$. The converse is however not true: observing a 
maximum memorization score of $M^{\text{LOO}}_i = \varepsilon$ for a 
particular model does not imply that the model is also $\varepsilon$-DP.  This 
connection of the memorization score to differential privacy offers additional 
support for the proposed formulation of the memorization score.

An alternative approach to limit memorization is to explicitly incorporate an 
outlier component in the model that would allow it to ignore atypical 
observations when learning the probability density. This technique has been 
previously used to handle outliers in factorial switching models 
\cite{quinn2009factorial} and to perform outlier detection in VAEs for tabular 
data \cite{eduardo2020robust}. The intuition is that by including a model 
component with broad support but low probability (such as a Gaussian with high 
variance), the log probability for atypical observations will be small whether 
they are included in the training data or not, resulting in a low memorization 
score. Other approaches such as using robust divergence measures instead of 
the KL-divergence in VAEs \cite{akrami2019robust} may also be able to 
alleviate memorization.

\section{Discussion}
\label{sec:discussion}

We have introduced a principled formulation of a memorization score for 
probabilistic generative models. The memorization score directly measures the 
impact of removing an observation on the model, and thereby allows us to 
quantify the degree to which the model has memorized it. We explored how the 
memorization score evolves during training and how it relates to typical 
nearest neighbor tests, and we have shown that highly memorized observations 
are not necessarily unlikely under the model. The proposed memorization score 
can be used to determine regions of the input space that require additional 
data collection, to understand the degree of memorization that an algorithm 
exhibits, or to identify training observations that must be pruned to avoid 
memorization by the model.

A question that requires further study is what constitutes a ``high'' 
memorization score on a particular data set. One of the main difficulties with 
this is that density estimates returned by a model, and thus probability 
differences, are not necessarily comparable between data sets 
\cite{nalisnick2019deep}. We expect that future work will focus on this 
important question, and suggest that inspiration may be taken from work on 
choosing $\varepsilon$ in differential privacy 
\cite{lee2011much,hsu2014differential}. Furthermore, exploring the 
relationship between memorization and the double descent phenomenon 
\cite{belkin2019reconciling,nakkiran2020deep} could be worthy of 
investigation. Improving the efficiency of the estimator is also considered an 
important topic for future research.

If we want diversity in the samples created by generative models, then the 
model will have to learn to generalize to regions of the data manifold that 
are not well represented in the input data. Whether this is achieved by 
extrapolating from other regions of the space or fails due to memorization is 
an important question. Our work thus contributes to the ongoing effort to 
understand the balance between memorization and generalization in deep 
generative neural networks. 

\begin{ack}
	The authors would like to thank the anonymous reviewers for their 
	helpful comments, and members of the AIDA project for helpful 
	discussions. This work was supported in part by The Alan Turing 
	Institute under EPSRC grant EP/N510129/1.
\end{ack}

\bibliographystyle{bibtex_style}
\bibliography{references}

\begin{thebibliography}{10}

\bibitem{kingma2014auto}
Kingma, D.~P. and Welling, M.
\newblock Auto-Encoding Variational {Bayes}.
\newblock In \emph{2nd International Conference on Learning Representations},
  edited by Y.~Bengio and Y.~LeCun, 2014.

\bibitem{rezende2014stochastic}
Rezende, D.~J., Mohamed, S., and Wierstra, D.
\newblock Stochastic Backpropagation and Approximate Inference in Deep
  Generative Models.
\newblock In \emph{Proceedings of the 31st International Conference on Machine
  Learning}, edited by E.~P. Xing and T.~Jebara, volume~32 of \emph{Proceedings
  of Machine Learning Research}, pp. 1278--1286. PMLR, Beijing, China, 2014.

\bibitem{goodfellow2014generative}
Goodfellow, I., Pouget-Abadie, J., Mirza, M., Xu, B., Warde-Farley, D., Ozair,
  S., Courville, A., and Bengio, Y.
\newblock Generative Adversarial Nets.
\newblock In \emph{Advances in Neural Information Processing Systems}, edited
  by Z.~Ghahramani, M.~Welling, C.~Cortes, N.~Lawrence, and K.~Q. Weinberger,
  volume~27. Curran Associates, Inc., 2014.

\bibitem{tabak2013family}
Tabak, E.~G. and Turner, C.~V.
\newblock A Family of Nonparametric Density Estimation Algorithms.
\newblock \emph{Communications on Pure and Applied Mathematics},
  66(2):145--164, 2013.

\bibitem{rezende2015variational}
Rezende, D. and Mohamed, S.
\newblock Variational Inference with Normalizing Flows.
\newblock In \emph{Proceedings of the 32nd International Conference on Machine
  Learning}, edited by F.~R. Bach and D.~Blei, volume~37 of \emph{Proceedings
  of Machine Learning Research}, pp. 1530--1538. PMLR, Lille, France, 2015.

\bibitem{sohl2015deep}
Sohl-Dickstein, J., Weiss, E., Maheswaranathan, N., and Ganguli, S.
\newblock Deep Unsupervised Learning using Nonequilibrium Thermodynamics.
\newblock In \emph{Proceedings of the 32nd International Conference on Machine
  Learning}, edited by F.~R. Bach and D.~Blei, volume~37 of \emph{Proceedings
  of Machine Learning Research}, pp. 2256--2265. PMLR, Lille, France, 2015.

\bibitem{ho2020denoising}
Ho, J., Jain, A., and Abbeel, P.
\newblock Denoising Diffusion Probabilistic Models.
\newblock In \emph{Advances in Neural Information Processing Systems}, edited
  by H.~Larochelle, M.~Ranzato, R.~Hadsell, M.~F. Balcan, and H.~Lin,
  volume~33. Curran Associates, Inc., 2020.

\bibitem{theis2016note}
Theis, L., van~den Oord, A., and Bethge, M.
\newblock A Note on the Evaluation of Generative Models.
\newblock In \emph{4th International Conference on Learning Representations},
  edited by Y.~Bengio and Y.~LeCun, 2016.

\bibitem{karras2018progressive}
Karras, T., Aila, T., Laine, S., and Lehtinen, J.
\newblock Progressive Growing of {GAN}s for Improved Quality, Stability, and
  Variation.
\newblock In \emph{6th International Conference on Learning Representations},
  2018.

\bibitem{brock2018large}
Brock, A., Donahue, J., and Simonyan, K.
\newblock Large Scale {GAN} Training for High Fidelity Natural Image Synthesis.
\newblock In \emph{7th International Conference on Learning Representations},
  2019.

\bibitem{vahdat2020nvae}
Vahdat, A. and Kautz, J.
\newblock {NVAE}: A Deep Hierarchical Variational Autoencoder.
\newblock In \emph{Advances in Neural Information Processing Systems}, edited
  by H.~Larochelle, M.~Ranzato, R.~Hadsell, M.~F. Balcan, and H.~Lin,
  volume~33, pp. 19667--19679. Curran Associates, Inc., 2020.

\bibitem{barz2020do}
Barz, B. and Denzler, J.
\newblock Do We Train on Test Data? {P}urging {CIFAR} of Near-Duplicates.
\newblock \emph{Journal of Imaging}, 6(6):41, 2020.

\bibitem{rosenblatt1956remarks}
Rosenblatt, M.
\newblock {Remarks on Some Nonparametric Estimates of a Density Function}.
\newblock \emph{The Annals of Mathematical Statistics}, 27(3):832 -- 837, 1956.

\bibitem{parzen1962estimation}
Parzen, E.
\newblock {On Estimation of a Probability Density Function and Mode}.
\newblock \emph{The Annals of Mathematical Statistics}, 33(3):1065 -- 1076,
  1962.

\bibitem{feldman2019does}
Feldman, V.
\newblock Does Learning Require Memorization? A Short Tale about a Long Tail.
\newblock \emph{arXiv preprint arXiv:1906.05271}, 2019.

\bibitem{feldman2020what}
Feldman, V. and Zhang, C.
\newblock What Neural Networks Memorize and Why: Discovering the Long Tail via
  Influence Estimation.
\newblock In \emph{Advances in Neural Information Processing Systems}, edited
  by H.~Larochelle, M.~Ranzato, R.~Hadsell, M.~F. Balcan, and H.~Lin,
  volume~33, pp. 2881--2891. Curran Associates, Inc., 2020.

\bibitem{zhang2017understanding}
Zhang, C., Bengio, S., Hardt, M., Recht, B., and Vinyals, O.
\newblock Understanding Deep Learning Requires Rethinking Generalization.
\newblock In \emph{5th International Conference on Learning Representations},
  2017.

\bibitem{arpit2017closer}
Arpit, D., Jastrz{\k{e}}bski, S., Ballas, N., Krueger, D., Bengio, E., Kanwal,
  M.~S., Maharaj, T., Fischer, A., Courville, A., Bengio, Y., and
  Lacoste-Julien, S.
\newblock A Closer Look at Memorization in Deep Networks.
\newblock In \emph{Proceedings of the 34th International Conference on Machine
  Learning}, edited by D.~Precup and Y.~W. Teh, volume~70, pp. 233--242, 2017.

\bibitem{stephenson2021on}
Stephenson, C., Padhy, S., Ganesh, A., Hui, Y., Tang, H., and Chung, S.
\newblock On the Geometry of Generalization and Memorization in Deep Neural
  Networks.
\newblock In \emph{9th International Conference on Learning Representations},
  2021.

\bibitem{carlini2019secret}
Carlini, N., Liu, C., Erlingsson, {\'U}., Kos, J., and Song, D.
\newblock The Secret Sharer: Evaluating and Testing Unintended Memorization in
  Neural Networks.
\newblock In \emph{28th {USENIX} Security Symposium}, pp. 267--284, 2019.

\bibitem{hochreiter1997long}
Hochreiter, S. and Schmidhuber, J.
\newblock Long Short-Term Memory.
\newblock \emph{Neural Computation}, 9(8):1735--1780, 1997.

\bibitem{jiang2021characterizing}
Jiang, Z., Zhang, C., Talwar, K., and Mozer, M.~C.
\newblock Characterizing Structural Regularities of Labeled Data in
  Overparameterized Models.
\newblock In \emph{Proceedings of the 38th International Conference on Machine
  Learning}, edited by M.~Meila and T.~Zhang, volume 139 of \emph{Proceedings
  of Machine Learning Research}, pp. 5034--5044. PMLR, 2021.

\bibitem{liu2020early}
Liu, S., Niles-Weed, J., Razavian, N., and Fernandez-Granda, C.
\newblock Early-Learning Regularization Prevents Memorization of Noisy Labels.
\newblock In \emph{Advances in Neural Information Processing Systems}, edited
  by H.~Larochelle, M.~Ranzato, R.~Hadsell, M.~F. Balcan, and H.~Lin,
  volume~33, pp. 20331--20342. Curran Associates, Inc., 2020.

\bibitem{xia2021robust}
Xia, X., Liu, T., Han, B., Gong, C., Wang, N., Ge, Z., and Chang, Y.
\newblock Robust Early-Learning: Hindering the Memorization of Noisy Labels.
\newblock In \emph{9th International Conference on Learning Representations},
  2021.

\bibitem{lopez2016revisiting}
Lopez-Paz, D. and Oquab, M.
\newblock Revisiting Classifier Two-Sample Tests.
\newblock In \emph{5th International Conference on Learning Representations},
  2017.

\bibitem{xu2018empirical}
Xu, Q., Huang, G., Yuan, Y., Guo, C., Sun, Y., Wu, F., and Weinberger, K.~Q.
\newblock An Empirical Study on Evaluation Metrics of Generative Adversarial
  Networks.
\newblock \emph{arXiv preprint arXiv:1806.07755}, 2018.

\bibitem{meehan2020non}
Meehan, C., Chaudhuri, K., and Dasgupta, S.
\newblock A Non-Parametric Test to Detect Data-Copying in Generative Models.
\newblock In \emph{Proceedings of the 23rd International Conference on
  Artificial Intelligence and Statistics}, edited by S.~Chiappa and
  R.~Calandra, volume 108 of \emph{Proceedings of Machine Learning Research},
  pp. 3546--3556. PMLR, 2020.

\bibitem{fredrikson2015model}
Fredrikson, M., Jha, S., and Ristenpart, T.
\newblock Model Inversion Attacks that Exploit Confidence Information and Basic
  Countermeasures.
\newblock In \emph{Proceedings of the 22nd {ACM} {SIGSAC} Conference on
  Computer and Communications Security}, pp. 1322--1333, 2015.

\bibitem{shokri2017membership}
Shokri, R., Stronati, M., Song, C., and Shmatikov, V.
\newblock Membership Inference Attacks against Machine Learning Models.
\newblock In \emph{2017 IEEE Symposium on Security and Privacy}, pp. 3--18.
  IEEE, 2017.

\bibitem{hitaj2017deep}
Hitaj, B., Ateniese, G., and Perez-Cruz, F.
\newblock Deep Models Under the {GAN}: Information Leakage from Collaborative
  Deep Learning.
\newblock In \emph{Proceedings of the 2017 {ACM} {SIGSAC} Conference on
  Computer and Communications Security}, pp. 603--618, 2017.

\bibitem{hayes2019logan}
Hayes, J., Melis, L., Danezis, G., and De~Cristofaro, E.
\newblock {LOGAN}: Membership Inference Attacks Against Generative Models.
\newblock \emph{Proceedings on Privacy Enhancing Technologies},
  2019(1):133--152, 2019.

\bibitem{webster2019detecting}
Webster, R., Rabin, J., Simon, L., and Jurie, F.
\newblock Detecting Overfitting of Deep Generative Networks via Latent
  Recovery.
\newblock In \emph{Proceedings of the {IEEE/CVF} Conference on Computer Vision
  and Pattern Recognition}, pp. 11273--11282, 2019.

\bibitem{arora2018do}
Arora, S., Risteski, A., and Zhang, Y.
\newblock Do {GAN}s Learn the Distribution? Some Theory and Empirics.
\newblock In \emph{6th International Conference on Learning Representations},
  2018.

\bibitem{nalisnick2019deep}
Nalisnick, E., Matsukawa, A., Teh, Y.~W., Gorur, D., and Lakshminarayanan, B.
\newblock Do Deep Generative Models Know What They Don't Know?
\newblock In \emph{7th International Conference on Learning Representations},
  2019.

\bibitem{salimans2016improved}
Salimans, T., Goodfellow, I., Zaremba, W., Cheun, V., Radford, A., and Chen, X.
\newblock Improved Techniques for Training {GAN}s.
\newblock In \emph{Advances in Neural Information Processing Systems}, edited
  by D.~D. Lee, M.~Sugiyama, U.~V. Luxburg, I.~Guyon, and R.~Garnett,
  volume~29, pp. 2234--2242. Curran Associates, Inc., 2016.

\bibitem{heusel2017gans}
Heusel, M., Ramsauer, H., Unterthiner, T., Nessler, B., and Hochreiter, S.
\newblock {GAN}s Trained by a Two Time-Scale Update Rule Converge to a Local
  {Nash} Equilibrium.
\newblock In \emph{Advances in Neural Information Processing Systems}, edited
  by I.~Guyon, U.~V. Luxburg, S.~Bengio, H.~Wallach, R.~Fergus,
  S.~Vishwanathan, and R.~Garnett, volume~30. Curran Associates, Inc., 2017.

\bibitem{gulrajani2018towards}
Gulrajani, I., Raffel, C., and Metz, L.
\newblock Towards {GAN} Benchmarks Which Require Generalization.
\newblock In \emph{7th International Conference on Learning Representations},
  2019.

\bibitem{hampel1974influence}
Hampel, F.~R.
\newblock The Influence Curve and Its Role in Robust Estimation.
\newblock \emph{Journal of the American Statistical Association},
  69(346):383--393, 1974.

\bibitem{cook1980characterizations}
Cook, R.~D. and Weisberg, S.
\newblock Characterizations of an Empirical Influence Function for Detecting
  Influential Cases in Regression.
\newblock \emph{Technometrics}, 22(4):495--508, 1980.

\bibitem{koh2017understanding}
Koh, P.~W. and Liang, P.
\newblock Understanding Black-box Predictions via Influence Functions.
\newblock In \emph{Proceedings of the 34th International Conference on Machine
  Learning}, edited by D.~Precup and Y.~W. Teh, volume~70, pp. 1885--1894,
  2017.

\bibitem{terashita2021influence}
Terashita, N., Ohashi, H., Nonaka, Y., and Kanemaru, T.
\newblock Influence Estimation for Generative Adversarial Networks.
\newblock In \emph{9th International Conference on Learning Representations},
  2021.

\bibitem{basu2021influence}
Basu, S., Pope, P., and Feizi, S.
\newblock Influence Functions in Deep Learning Are Fragile.
\newblock In \emph{9th International Conference on Learning Representations},
  2021.

\bibitem{kong2021understanding}
Kong, Z. and Chaudhuri, K.
\newblock Understanding Instance-based Interpretability of Variational
  Auto-Encoders.
\newblock In \emph{Advances in Neural Information Processing Systems},
  volume~34, 2021.

\bibitem{pruthi2020estimating}
Pruthi, G., Liu, F., Kale, S., and Sundararajan, M.
\newblock Estimating Training Data Influence by Tracing Gradient Descent.
\newblock In \emph{Advances in Neural Information Processing Systems}, edited
  by H.~Larochelle, M.~Ranzato, R.~Hadsell, M.~F. Balcan, and H.~Lin,
  volume~33, pp. 19920--19930. Curran Associates, Inc., 2020.

\bibitem{bousquet2002stability}
Bousquet, O. and Elisseeff, A.
\newblock Stability and Generalization.
\newblock \emph{Journal of Machine Learning Research}, 2:499--526, 2002.

\bibitem{kullback1951information}
Kullback, S. and Leibler, R.~A.
\newblock On Information and Sufficiency.
\newblock \emph{The Annals of Mathematical Statistics}, 22(1):79--86, 1951.

\bibitem{burda2016importance}
Burda, Y., Grosse, R.~B., and Salakhutdinov, R.
\newblock Importance Weighted Autoencoders.
\newblock In \emph{4th International Conference on Learning Representations},
  edited by Y.~Bengio and Y.~LeCun, 2016.

\bibitem{lecun1998mnist}
LeCun, Y., Cortes, C., and Burges, C. J.~C.
\newblock The {MNIST} Database of Handwritten Digits, 1998.

\bibitem{krizhevsky2009learning}
Krizhevsky, A.
\newblock \emph{Learning Multiple Layers of Features from Tiny Images}.
\newblock Master's thesis, University of Toronto, 2009.

\bibitem{liu2015faceattributes}
Liu, Z., Luo, P., Wang, X., and Tang, X.
\newblock Deep Learning Face Attributes in the Wild.
\newblock In \emph{Proceedings of International Conference on Computer Vision},
  2015.

\bibitem{kingma2015adam}
Kingma, D.~P. and Ba, J.
\newblock Adam: {A} Method for Stochastic Optimization.
\newblock In \emph{3rd International Conference on Learning Representations},
  edited by Y.~Bengio and Y.~LeCun, 2015.

\bibitem{paszke2019pytorch}
Paszke, A., Gross, S., Massa, F., Lerer, A., Bradbury, J., Chanan, G., Killeen,
  T., Lin, Z., Gimelshein, N., Antiga, L., Desmaison, A., K{\"{o}}pf, A., Yang,
  E., DeVito, Z., Raison, M., Tejani, A., Chilamkurthy, S., Steiner, B., Fang,
  L., Bai, J., and Chintala, S.
\newblock {PyTorch}: An Imperative Style, High-Performance Deep Learning
  Library.
\newblock In \emph{Advances in Neural Information Processing Systems}, edited
  by H.~Wallach, H.~Larochelle, A.~Beygelzimer, F.~d'Alch{\'e} Buc, E.~B. Fox,
  and R.~Garnett, volume~32. Curran Associates, Inc., 2019.

\bibitem{wu2016quantitative}
Wu, Y., Burda, Y., Salakhutdinov, R., and Grosse, R.~B.
\newblock On the Quantitative Analysis of Decoder-Based Generative Models.
\newblock In \emph{5th International Conference on Learning Representations},
  2017.

\bibitem{dwork2006calibrating}
Dwork, C., McSherry, F., Nissim, K., and Smith, A.
\newblock Calibrating Noise to Sensitivity in Private Data Analysis.
\newblock In \emph{Theory of Cryptography Conference}, pp. 265--284. Springer,
  2006.

\bibitem{dwork2014algorithmic}
Dwork, C. and Roth, A.
\newblock The Algorithmic Foundations of Differential Privacy.
\newblock \emph{Foundations and Trends in Theoretical Computer Science},
  9(3–4):211--407, 2014.

\bibitem{quinn2009factorial}
Quinn, J.~A., Williams, C. K.~I., and McIntosh, N.
\newblock Factorial Switching Linear Dynamical Systems Applied to Physiological
  Condition Monitoring.
\newblock \emph{{IEEE} Transactions on Pattern Analysis and Machine
  Intelligence}, 31(9):1537--1551, 2009.

\bibitem{eduardo2020robust}
Eduardo, S., Naz{\'a}bal, A., Williams, C. K.~I., and Sutton, C.
\newblock Robust Variational Autoencoders for Outlier Detection and Repair of
  Mixed-Type Data.
\newblock In \emph{International Conference on Artificial Intelligence and
  Statistics}, pp. 4056--4066. PMLR, 2020.

\bibitem{akrami2019robust}
Akrami, H., Joshi, A.~A., Li, J., Aydore, S., and Leahy, R.~M.
\newblock Robust Variational Autoencoder.
\newblock \emph{arXiv preprint arXiv:1905.09961}, 2019.

\bibitem{lee2011much}
Lee, J. and Clifton, C.
\newblock How Much is Enough? Choosing $\varepsilon$ for Differential Privacy.
\newblock In \emph{International Conference on Information Security}, pp.
  325--340. Springer, 2011.

\bibitem{hsu2014differential}
Hsu, J., Gaboardi, M., Haeberlen, A., Khanna, S., Narayan, A., Pierce, B.~C.,
  and Roth, A.
\newblock Differential Privacy: An Economic Method for Choosing Epsilon.
\newblock In \emph{2014 IEEE 27th Computer Security Foundations Symposium}, pp.
  398--410. IEEE, 2014.

\bibitem{belkin2019reconciling}
Belkin, M., Hsu, D., Ma, S., and Mandal, S.
\newblock Reconciling Modern Machine-Learning Practice and the Classical
  Bias--Variance Trade-off.
\newblock \emph{Proceedings of the National Academy of Sciences},
  116(32):15849--15854, 2019.

\bibitem{nakkiran2020deep}
Nakkiran, P., Kaplun, G., Bansal, Y., Yang, T., Barak, B., and Sutskever, I.
\newblock Deep Double Descent: Where Bigger Models and More Data Hurt.
\newblock In \emph{8th International Conference on Learning Representations},
  2020.

\bibitem{song2021scorebased}
Song, Y., Sohl-Dickstein, J., Kingma, D.~P., Kumar, A., Ermon, S., and Poole,
  B.
\newblock Score-Based Generative Modeling through Stochastic Differential
  Equations.
\newblock In \emph{9th International Conference on Learning Representations},
  2021.

\bibitem{salakhutdinov2008quantitative}
Salakhutdinov, R. and Murray, I.
\newblock On the Quantitative Analysis of Deep Belief Networks.
\newblock In \emph{Proceedings of the 25th International Conference on Machine
  Learning}, edited by A.~McCallum and S.~Roweis, pp. 872--879. Omnipress,
  2008.

\bibitem{nair2010rectified}
Nair, V. and Hinton, G.~E.
\newblock Rectified Linear Units Improve Restricted {Boltzmann} Machines.
\newblock In \emph{Proceedings of the 27th International Conference on Machine
  Learning}, edited by J.~F{\"u}rnkranz and T.~Joachims, pp. 807--814.
  Omnipress, 2010.

\bibitem{uria2013rnade}
Uria, B., Murray, I., and Larochelle, H.
\newblock {RNADE}: The Real-Valued Neural Autoregressive Density-Estimator.
\newblock In \emph{Advances in Neural Information Processing Systems}, edited
  by C.~J.~C. Burges, L.~Bottou, M.~Welling, Z.~Ghahramani, and K.~Q.
  Weinberger, volume~26, pp. 2175--2183. Curran Associates, Inc., 2013.

\bibitem{papamakarios2017masked}
Papamakarios, G., Pavlakou, T., and Murray, I.
\newblock Masked Autoregressive Flow for Density Estimation.
\newblock In \emph{Advances in Neural Information Processing Systems}, edited
  by I.~Guyon, U.~V. Luxburg, S.~Bengio, H.~Wallach, R.~Fergus,
  S.~Vishwanathan, and R.~Garnett, volume~30. Curran Associates, Inc., 2017.

\bibitem{radford2016unsupervised}
Radford, A., Metz, L., and Chintala, S.
\newblock Unsupervised Representation Learning with Deep Convolutional
  Generative Adversarial Networks.
\newblock In \emph{4th International Conference on Learning Representations},
  edited by Y.~Bengio and Y.~LeCun, 2016.

\bibitem{ioffe2015batch}
Ioffe, S. and Szegedy, C.
\newblock Batch Normalization: Accelerating Deep Network Training by Reducing
  Internal Covariate Shift.
\newblock In \emph{Proceedings of the 32nd International Conference on Machine
  Learning}, edited by F.~R. Bach and D.~Blei, volume~37 of \emph{Proceedings
  of Machine Learning Research}, pp. 448--456. PMLR, Lille, France, 2015.

\bibitem{maas2013rectifier}
Maas, A.~L., Hannun, A.~Y., and Ng, A.~Y.
\newblock Rectifier Nonlinearities Improve Neural Network Acoustic Models.
\newblock In \emph{Proceedings of the 30th International Conference on Machine
  Learning}, edited by S.~Dasgupta and D.~McAllester, volume~28 of
  \emph{Proceedings of Machine Learning Research}. PMLR, 2013.

\end{thebibliography}

\clearpage
\appendix

\begin{center}
{\large\bfseries On Memorization in Probabilistic Deep Generative 
	Models\\[.1em]
	Supplementary Material}
\end{center}

\section{Memorized observations in recently proposed generative models}
\label{app:copies}

\begin{wrapfigure}{r}{.3\textwidth}
	\vskip-.5\baselineskip
	\centering%
	\captionsetup[subfigure]{justification=centering}%
	\begin{subfigure}[b]{.10\textwidth}
		\centering
		\includegraphics[width=.9\textwidth]{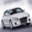}
		\caption{\label{fig:white_audi}}
	\end{subfigure}
	\quad
	\begin{subfigure}[b]{.10\textwidth}
		\centering
		\includegraphics[width=.9\textwidth]{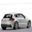}
		\caption{\label{fig:red_taillight}}
	\end{subfigure}
	\caption{Examples of images from the CIFAR-10 training set that were 
		spotted in illustrations of samples from the model in recent 
		work on generative models.}%
	\label{fig:autos}
\end{wrapfigure}

While experimenting with the proposed memorization score on CIFAR-10 
\cite{krizhevsky2009learning}, we noticed that the images of automobiles shown 
in Figure~\ref{fig:autos} are present in the training set multiple times (with 
slight variation). We subsequently spotted these images in the illustrations 
of generated samples in \cite{ho2020denoising} (Figure 13, example (a) can be 
seen twice) and \cite{song2021scorebased} (Figure 11 and Figure 13, truck 
class). These works are recently proposed probabilistic generative models that 
achieve impressive performance on sample quality metrics such as the inception 
score (IS) \cite{salimans2016improved} and the Fr\'echet inception distance 
(FID) \cite{heusel2017gans}, and also achieve high log likelihoods.  However, 
the fact that we were able to serendipitously spot images from the training 
set in the generated samples might suggest that some unintended memorization 
occurs in these models. We do not know if there are other images in the 
presented samples that are present in the training data. Of course, spotting 
near duplicates of training observations is only possible because these models 
yield realistic samples. As we argue in the main text and as has been shown by 
previous works \cite{gulrajani2018towards,webster2019detecting}, quality 
metrics such as IS and FID do not detect memorization.

We emphasize that this evidence is presented mainly to support the notion that 
(unintended) memorization can occur in probabilistic deep generative models, 
and to provide additional motivation for understanding and quantifying when 
and how memorization arises, which is the focus of our work.

\section{Experimental Details}
\label{app:details}

This section describes additional details of the data sets, model 
architectures, and experimental setup.

\subsection{Datasets}%
\label{sub:datasets}

We use the MNIST \cite{lecun1998mnist}, CIFAR-10 
\cite{krizhevsky2009learning}, and CelebA \cite{liu2015faceattributes} data 
sets, which are widely used and are freely available for research purposes 
(although to the best of our knowledge explicit licenses are not available). 
For MNIST we binarize the images dynamically during training by considering 
each grayscale pixel value as the parameter of an independent Bernoulli 
variable, as is common \cite{salakhutdinov2008quantitative,vahdat2020nvae}.  
Images in all data sets are resized to $32\times 32$ pixels for efficiency and 
ease of implementation. CIFAR-10 contains color images from 10 different 
categories and does not require further preprocessing. CelebA contains 
potentially identifiable images of faces of celebrities sourced from publicly 
available images on the Internet. We used the predefined cropping function of 
\cite{vahdat2020nvae} to center the face region. For CIFAR-10 and CelebA we 
used random horizontal flips during training as data augmentation. All data 
sets have predefined train and test sets, and CelebA additionally has a 
validation set. We mainly used the training sets in the experiments, with the 
exception of the experiments for Figure~\ref{fig:mnist_mem_lrs_loss}, which 
uses the MNIST test set, and the experiments in 
Section~\ref{sub:nearest_neighbors}, which use the CelebA validation set.

\subsection{Model Architectures}%
\label{sub:model_architectures}

Let $L$ denote the size of the latent space and recall that $\bft{x} \in 
\mathcal{X} \subseteq \mathbb{R}^D$. For all experiments we used a Gaussian 
encoder with a learned diagonal covariance matrix, $q_{\phi}(\bft{z} \given 
\bft{x}) = \mathcal{N}(\bft{z} ; \bfs{\mu}_{\phi}(\bft{x}), 
\text{diag}(\bfs{\sigma}_{\phi}^2(\bft{x})))$  and a standard multivariate 
Gaussian prior on the latent variables, $p(\bft{z}) = \mathcal{N}(\bft{z} ; 0, 
\bft{I}_L)$. As mentioned above we used a dynamically binarized version of the 
MNIST data set, and therefore used a Bernoulli likelihood for the decoder of 
the VAE. Both the encoder and decoder used fully connected layers with the 
\textsc{ReLU} activation on the intermediate layers \cite{nair2010rectified} 
and a sigmoid activation on the output of the decoder that represents the 
parameter of the Bernoulli distribution. For MNIST we used $L = 16$. Full 
details of the model architecture are given in Table~\ref{tab:architectures}.

\begin{table}[tb]
	\centering
	\caption{Model architectures used for the experiments. We used fully 
		connected (\textsc{FC}) layers with the \textsc{ReLU} 
		activation for MNIST, with the \textsc{Sigmoid} activation on 
		the decoder. For CIFAR-10 and CelebA we used convolutional 
		layers for the encoder (\textsc{Conv2d} with kernel size~4, 
		stride~2, and padding~1), followed by batch normalization 
		(\textsc{BN}), and the Leaky ReLU activation (\textsc{LReLU}, 
		using slope~0.2). For these data sets the decoder consists of 
		transposed convolution layers (\textsc{ConvT2d}, with kernel 
		size~4, stride~2, and padding~1 except for the layer marked 
		with an asterisk (*), which uses kernel size~2, stride~1, and 
		padding~0 to get the correct output size), followed by batch 
		norm and the ReLU activation.  We use the abbreviations 
		$\textsc{EncBlock}(C_1, C_2) = \textsc{Conv2d}(C_1, C_2) 
		\rightarrow \textsc{BN} \rightarrow \textsc{LReLU}$ and 
		$\textsc{DecBlock}(C_1, C_2) = \textsc{ConvT2d}(C_1, C_2) 
		\rightarrow \textsc{BN} \rightarrow \textsc{ReLU}$.}
	\label{tab:architectures}
	\def\RA{$\rightarrow\,$}
	\small
	\begin{tabular}{lrrr}
		\toprule
		Data set & Encoder network & Decoder network & Likelihood ($p_{\theta}(\bft{x} \given \bft{z})$) \\
		\midrule
		MNIST & $\textsc{FC}(1024, 512)$ \RA \textsc{ReLU} & $\textsc{FC}(L, 256)$ \RA ReLU & $\mathcal{B}(x_{ij}; \pi_{ij}(\bft{z}))$ \\
		      & \RA $\textsc{FC}(512, 256)$ \RA \textsc{ReLU} & \RA $\textsc{FC}(256, 512)$ \RA ReLU & \\
		      & \RA $\textsc{FC}(256, L)$, $\textsc{FC}(256, L)$ & \RA $\textsc{FC}(512, 1024)$ & \\
		      &                                                  & \RA \textsc{Sigmoid} & \\
		& & & \\
		CIFAR-10 & $\textsc{Conv2d}(C, F)$ \RA $\textsc{LReLU}$ & 
		$\textsc{DecBlock}^*(L, 8F)$ & $\mathcal{N}(\bft{x} ; 
		\bfs{\mu}_{\theta}(\bft{z}), 
		\text{diag}(\bfs{\sigma}_{\theta}(\bft{z})))$  \\
			 & \RA $\textsc{EncBlock}(F,  2F)$  & \RA $\textsc{DecBlock}(8F, 4F)$ & \\
			 & \RA $\textsc{EncBlock}(2F, 4F)$  & \RA $\textsc{DecBlock}(4F, 2F)$ & \\
			 & \RA $\textsc{EncBlock}(4F, 8F)$  & \RA $\textsc{DecBlock}(2F, F)$ & \\
			 & \RA \textsc{Flatten}  & \RA $\textsc{ConvT2d}(F, 2C)$ & \\
			 & \RA $\textsc{FC}(32F, L)$, $\textsc{FC}(32F, L)$ & & \\
		& & & \\
		CelebA & Same as for CIFAR-10 & Same as for CIFAR-10, & $\mathcal{N}(\bft{x} ; \bfs{\mu}_{\theta}(\bft{z}), \gamma_{\theta} \bft{I}_D)$ \\
		& & except final layer uses & \\
		& & $\textsc{ConvT2d}(F, C)$ & \\
		\bottomrule
	\end{tabular}
\end{table}

For CIFAR-10 and CelebA we used a Gaussian likelihood for the decoder, 
employed uniform dequantization on the pixel values \cite{uria2013rnade}, and 
trained the models in logit space following \cite{papamakarios2017masked}.  
For both data sets we used an architecture similar to DCGAN 
\cite{radford2016unsupervised}, consisting of four convolutional layers in the 
encoder, each followed by batch normalization \cite{ioffe2015batch} and leaky 
\textsc{ReLU} activation \cite{maas2013rectifier}, and five transposed 
convolution layers in the decoder followed by batch normalization and 
\textsc{ReLU}, see Table~\ref{tab:architectures}. For CIFAR-10 the Gaussian 
likelihood on the decoder was parameterized as $p_{\theta}(\bft{x} \given 
\bft{z}) = \mathcal{N}(\bft{x} ; \bfs{\mu}_{\theta}(\bft{z}), 
\textrm{diag}(\bfs{\sigma}_{\theta}(\bft{z})))$ and for CelebA we used the 
simpler formulation $p_{\theta}(\bft{x} \given \bft{z}) = \mathcal{N}(\bft{x} 
; \bfs{\mu}_{\theta}(\bft{z}), \gamma_{\theta} \bft{I}_D)$ with a learned 
parameter $\gamma_{\theta}$, as the more general decoder was unnecessary. For 
CIFAR-10 and CelebA the number of input channels is $C = 3$ and we used $L = 
64$ and $L = 32$, respectively. For the convolutional networks the feature map 
multiplier was set to $F = 32$ (see Table~\ref{tab:architectures}).

\subsection{Training details}%
\label{sub:training_details}

We used Adam \cite{kingma2015adam} to optimize the parameters of the model 
with learning rate $\eta = 10^{-3}$ for the main experiments and $\eta = 
10^{-4}$ for the experiments on MNIST in Section~\ref{sub:results}. We used a 
batch size of 64 and left the remaining parameters for Adam at their default 
values in PyTorch \cite{paszke2019pytorch}. For both MNIST and CIFAR-10 we 
trained for 100 epochs, and used 50 epochs for CelebA. These settings were 
chosen by taking into consideration the available computational resources and 
aimed to avoid overtraining. The parameter settings were determined through 
some preliminary experimentation and were not extensively optimized.  
Experiments were conducted on a desktop machine running Arch Linux, using an 
NVIDIA GeForce GTX 1660 SUPER GPU, 32GB of RAM, and an AMD Ryzen 5 3600 
processor. Total wall-clock time was about 200 hours for the main results, 
excluding preliminary experimentation. Electricity needed for the experiments 
came from carbon-free sources.

%

As mentioned in the main text, importance sampling was used to approximate 
$p(\bft{x})$, such that
\begin{equation}
	p(\bft{x}) \approx \frac{1}{N} \sum_{l=1}^N \frac{%
		p_{\theta}(\bft{x} \given \bft{z}_l) p(\bft{z}_l)%
	}{%
		q_{\phi}(\bft{z}_l \given \bft{x})
	}, \qquad \bft{z}_l \sim q_{\phi}(\bft{z} \given \bft{x}).
\end{equation}
This was computed in log space for numerical accuracy. For MNIST we used $N = 
256$ and for CIFAR-10 and CelebA we used $N = 128$ samples.

\FloatBarrier
\section{Additional Results}%
\label{app:additional_results}

Below we show additional results that confirm the findings presented in the 
main text for different data sets.

\subsection{Qualitative Illustrations}%
\label{sub:qualitative_illustrations}

\begin{figure}[h]
	\centering
	\captionsetup[subfigure]{justification=centering}%
	\begin{subfigure}[b]{.30\textwidth}
		\includegraphics[width=.9\textwidth]{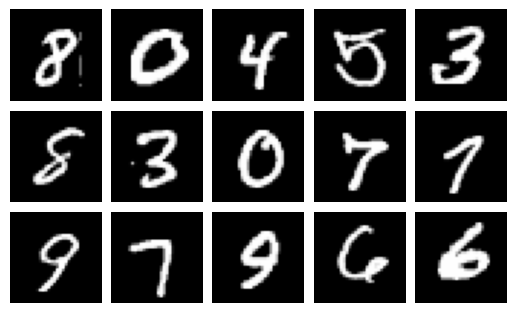}
		\caption{Low memorization \label{fig:app_mnist_lr3_bottom}}
	\end{subfigure}
	\quad
	\begin{subfigure}[b]{.30\textwidth}
		\includegraphics[width=.9\textwidth]{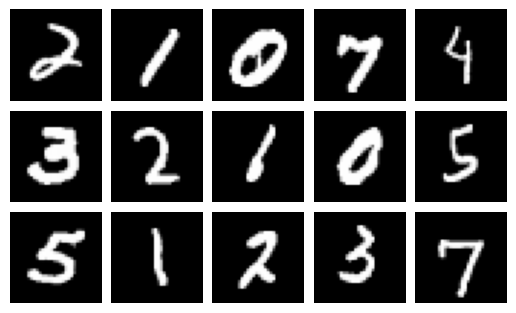}
		\caption{Median memorization \label{fig:app_mnist_lr3_middle}}
	\end{subfigure}
	\quad
	\begin{subfigure}[b]{.30\textwidth}
		\includegraphics[width=.9\textwidth]{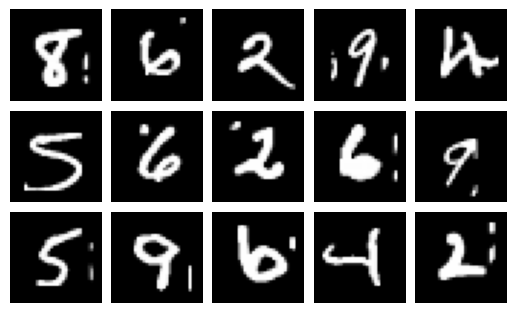}
		\caption{High memorization \label{fig:app_mnist_lr3_top}}
	\end{subfigure}
	\caption{Observations with low, median, and high memorization scores 
		in the MNIST data set, for a VAE trained using learning rate 
		$\eta = 10^{-3}$. Memorization scores range from about $-18$ 
		in the top left of figure (a) to about $200$ in the bottom 
		right of figure (c), with a median of $4.4$.  
		\label{fig:app_mnist_lr3_mem}}
\end{figure}

\begin{figure}[h]
	\centering
	\captionsetup[subfigure]{justification=centering}%
	\begin{subfigure}[b]{.30\textwidth}
		\includegraphics[width=.9\textwidth]{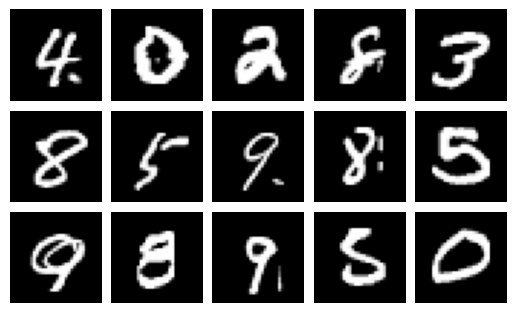}
		\caption{Low memorization \label{fig:app_mnist_lr4_bottom}}
	\end{subfigure}
	\quad
	\begin{subfigure}[b]{.30\textwidth}
		\includegraphics[width=.9\textwidth]{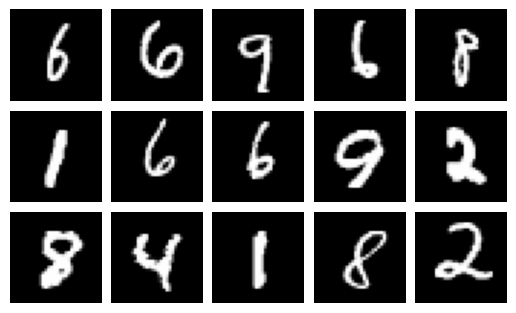}
		\caption{Median memorization \label{fig:app_mnist_lr4_middle}}
	\end{subfigure}
	\quad
	\begin{subfigure}[b]{.30\textwidth}
		\includegraphics[width=.9\textwidth]{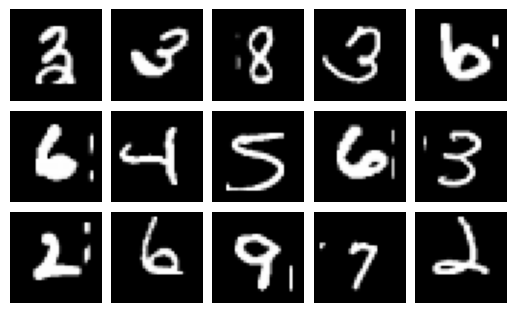}
		\caption{High memorization \label{fig:app_mnist_lr4_top}}
	\end{subfigure}
	\caption{Observations with low, median, and high memorization scores 
		in the MNIST data set, for a VAE trained using learning rate 
		$\eta = 10^{-4}$. Memorization scores range from about $-13$ 
		in the top left of figure (a) to about $80$ in the bottom 
		right of figure (c), with a median of $3.5$.  
		\label{fig:app_mnist_lr4_mem}}
\end{figure}

\begin{figure}[h]
	\centering
	\captionsetup[subfigure]{justification=centering}%
	\begin{subfigure}[b]{.30\textwidth}
		\includegraphics[width=.9\textwidth]{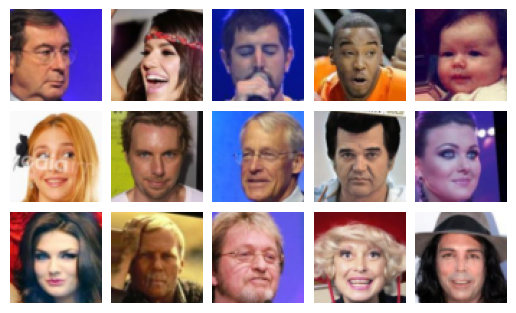}
		\caption{Low memorization \label{fig:app_celeba_bottom}}
	\end{subfigure}
	\quad
	\begin{subfigure}[b]{.30\textwidth}
		\includegraphics[width=.9\textwidth]{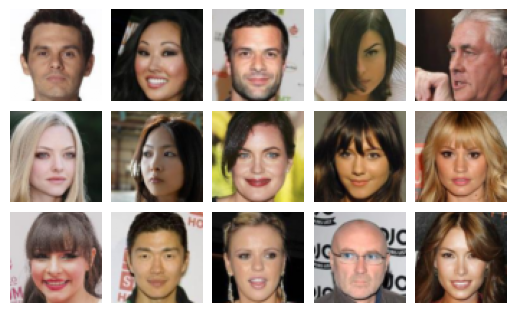}
		\caption{Median memorization \label{fig:app_celeba_middle}}
	\end{subfigure}
	\quad
	\begin{subfigure}[b]{.30\textwidth}
		\includegraphics[width=.9\textwidth]{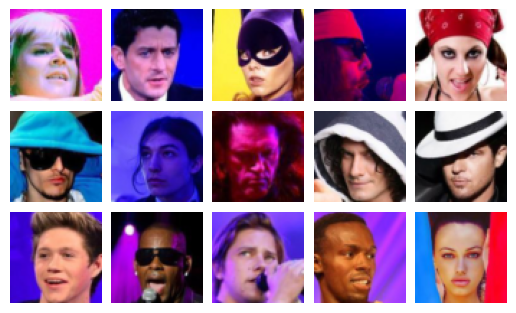}
		\caption{High memorization \label{fig:app_celeba_top}}
	\end{subfigure}
	\caption{Observations with low, median, and high memorization scores 
		in the CelebA data set when the density is learned using a 
		convolutional VAE.  Memorization scores range from about 
		$-450$ in the top left of figure (a) to about $6500$ in the 
		bottom right of figure (c), with a median of about $60$.  
		\label{fig:app_celeba_mem}}
\end{figure}

In Figures~\ref{fig:app_mnist_lr3_mem}, \ref{fig:app_mnist_lr4_mem}, and 
\ref{fig:app_celeba_mem} we illustrate observations with low, median, and high 
memorization scores for a VAE trained on MNIST using $\eta = 10^{-3}$, MNIST 
using $\eta = 10^{-4}$, and CelebA, respectively. As can be seen from the 
figures and as discussed in the main text in Section~\ref{sub:results}, while 
some of the highly memorized observations have visual anomalies, others are 
not unlike those that receive low memorization scores. For instance, for the 
VAE trained on MNIST with learning rate $\eta = 10^{-4}$, we see that images 
from both the low and high memorization groups have active pixels that are not 
part of the digit (compare, for instance, the images of 9s on the middle of 
the bottom rows of Figure~\ref{fig:app_mnist_lr4_bottom} and 
Figure~\ref{fig:app_mnist_lr4_top}).  

\FloatBarrier
\clearpage
\subsection{Outliers vs. Memorization}%
\label{sub:outliers_vs_memorization}

\begin{figure}[h]
	\centering
	\captionsetup[subfigure]{justification=centering}%
	\begin{subfigure}[b]{0.51\textwidth}
		\includegraphics[height=48mm]{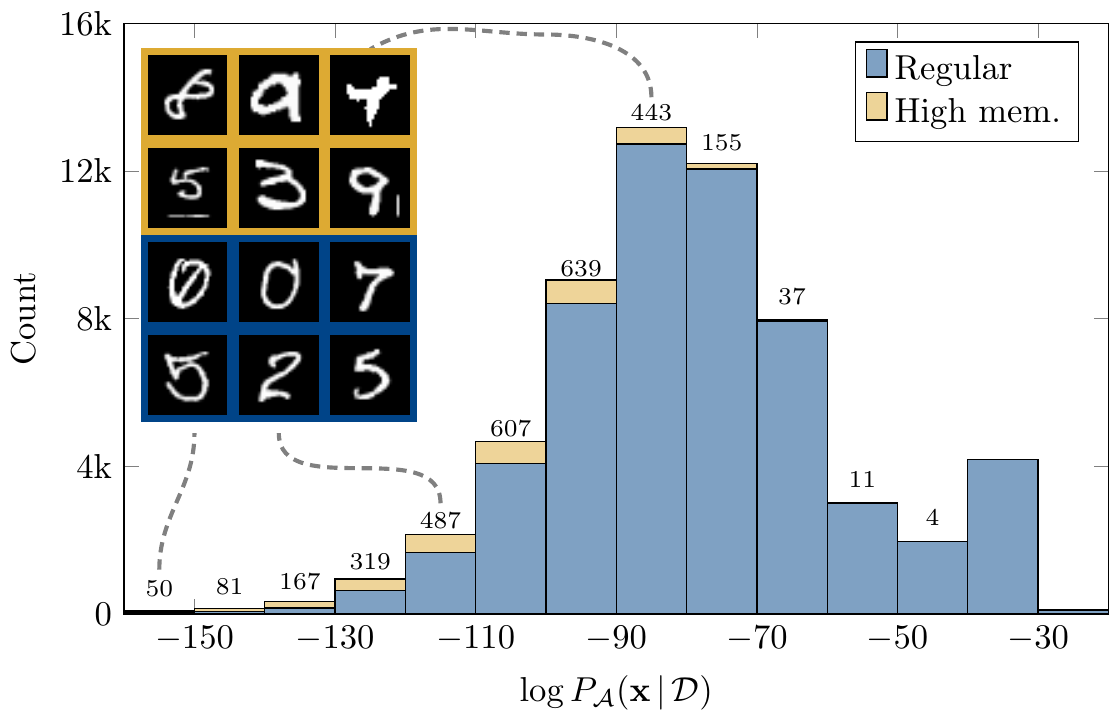}
		\caption{Counts \label{fig:app_mnist_lr3_logpx_bins}}
	\end{subfigure}
	\qquad
	\begin{subfigure}[b]{0.43\textwidth}
		\includegraphics[height=48mm]{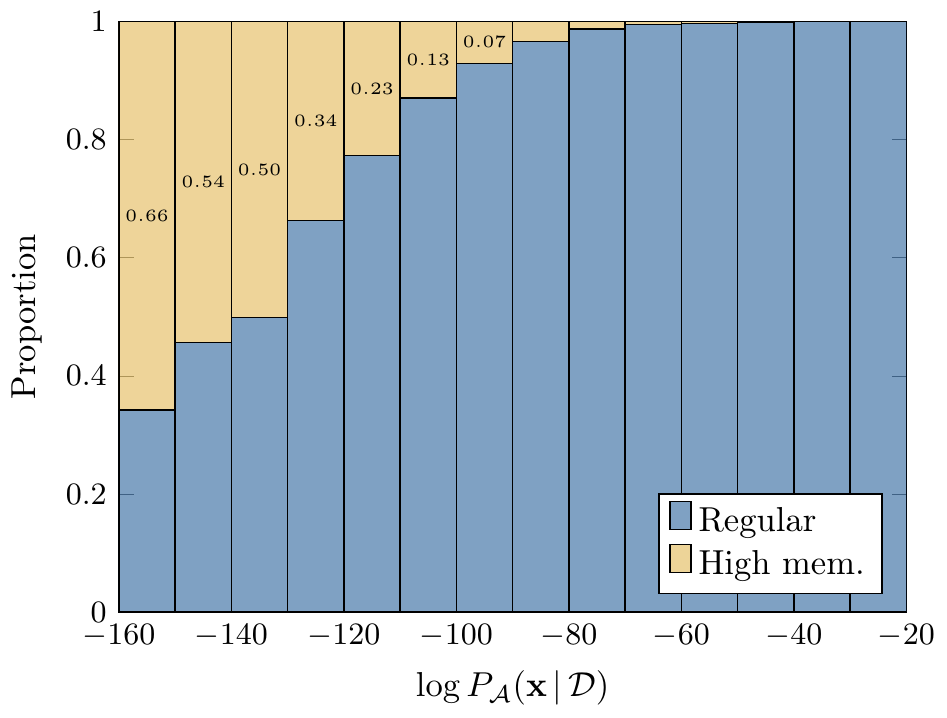}
		\caption{Proportions \label{fig:app_mnist_lr3_logpx_prop}}
	\end{subfigure}
	\caption{In (a) we show a histogram of the number of highly memorized 
		(\textcolor{MyYellow}{yellow}) and regular 
		(\textcolor{MyBlue}{blue}) observations for bins of the log 
		probability under a VAE model trained on the MNIST data set 
		using learning rate $\eta = 10^{-3}$. The numbers above the 
		bars correspond to the number of highly memorized observations 
		in each bin (for MNIST, $n = 60,000$). Randomly selected 
		training observations from several bins are shown, with dashed 
		lines illustrating the bin where the images in a particular 
		column can be found. Images with a yellow frame are highly 
		memorized whereas those with a blue frame have low 
		memorization scores. Figure~(b) shows the \emph{proportion} of 
		highly memorized and regular observations for each bin.
		\label{fig:app_mnist_lr3_logpxs}}
\end{figure}

\begin{figure}[h]
	\centering
	\captionsetup[subfigure]{justification=centering}%
	\begin{subfigure}[b]{0.51\textwidth}
		\includegraphics[height=48mm]{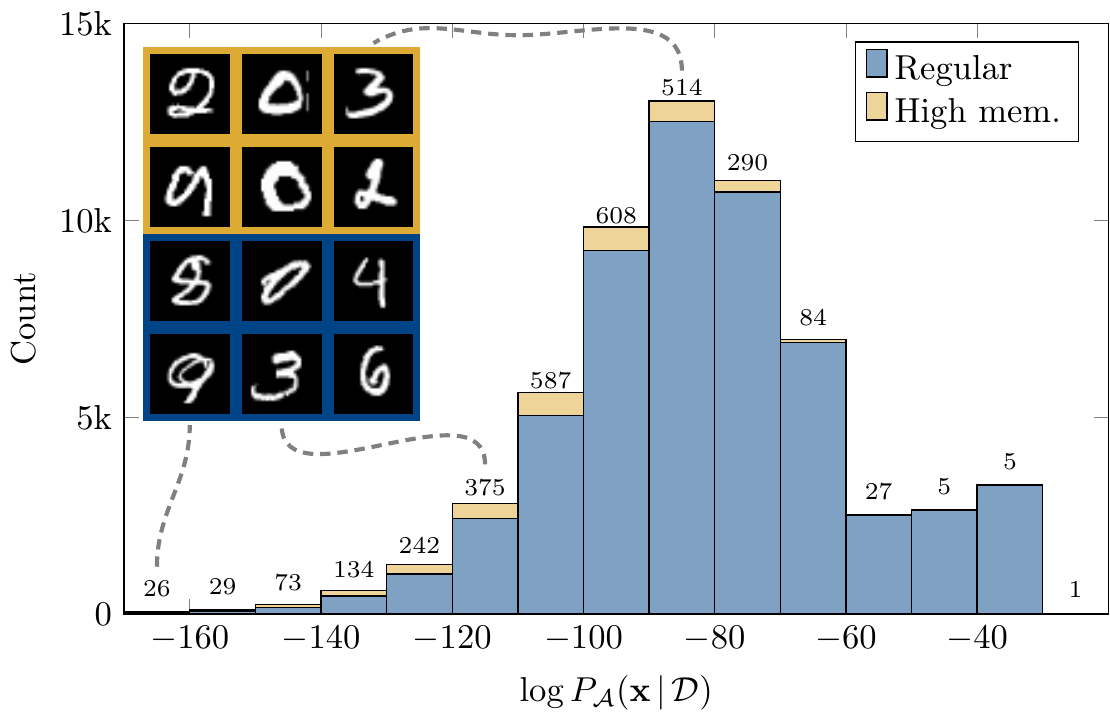}
		\caption{Counts \label{fig:app_mnist_lr4_logpx_bins}}
	\end{subfigure}
	\qquad
	\begin{subfigure}[b]{0.43\textwidth}
		\includegraphics[height=48mm]{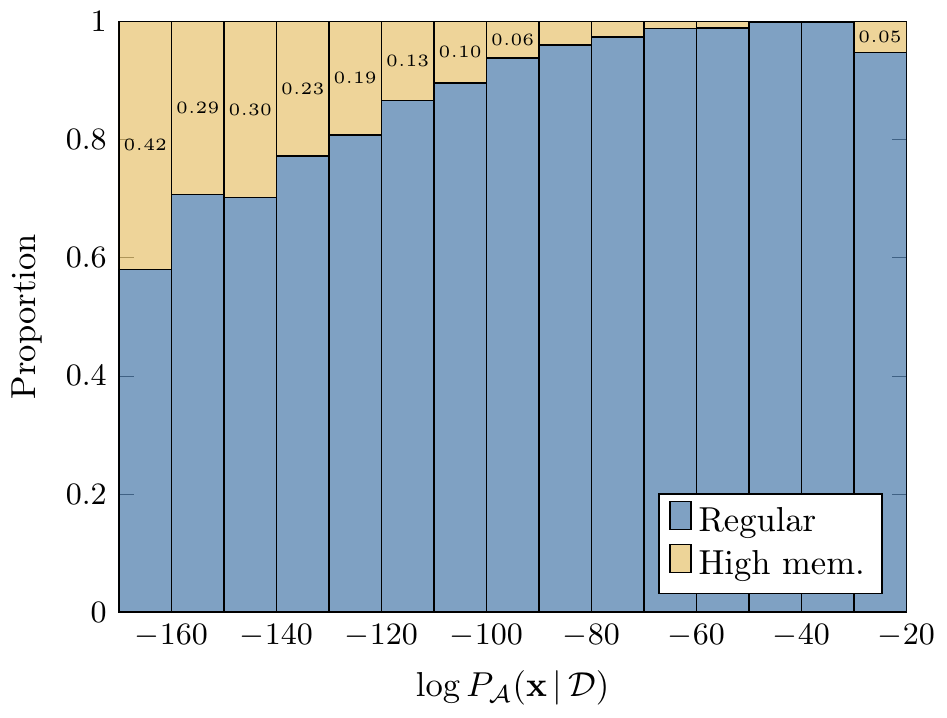}
		\caption{Proportions \label{fig:app_mnist_lr4_logpx_prop}}
	\end{subfigure}
	\caption{Similar to Figure~\ref{fig:app_mnist_lr3_logpxs}, but for the 
		VAE trained on MNIST with learning rate $\eta = 10^{-4}$.
		\label{fig:app_mnist_lr4_logpxs}}
\end{figure}

Figures~\ref{fig:app_mnist_lr3_logpxs} and \ref{fig:app_mnist_lr4_logpxs} 
replicate the experiments shown in Figure~\ref{fig:celeba_mem} in the main 
text for the VAE trained on the MNIST data set using two different learning 
rates. We again see that relatively high memorization is not exclusive to 
observations that receive a low probability under the model. Note that for 
this particular data set the density estimated by the VAE is slightly 
multimodal, with the peak in density for higher values of $\log 
P_{\mathcal{A}}(\bft{x} \given \mathcal{D})$ corresponding to observations for 
digit \verb+1+.

\FloatBarrier
\clearpage
\subsection{Nearest Neighbors}%
\label{app:nearest_neighbors}

The nearest neighbor experiments demonstrated in 
Section~\ref{sub:nearest_neighbors} are repeated below in 
Figures~\ref{fig:nn_vs_mem_mnist_lr3}~and~\ref{fig:nn_vs_mem_mnist_lr4} for 
the VAEs trained on the MNIST data set using learning rates of $10^{-3}$ and 
$10^{-4}$. For these models and data set we again do not see a clear relation 
between the nearest neighbor distance ratio $\rho_i$ and the proposed 
memorization score $M^{\text{K-fold}}_i$, as discussed in the main text.

\begin{figure}[h]
	\centering
	\captionsetup[subfigure]{justification=centering}%
	\begin{subfigure}[b]{.40\textwidth}
		\includegraphics[height=40mm]{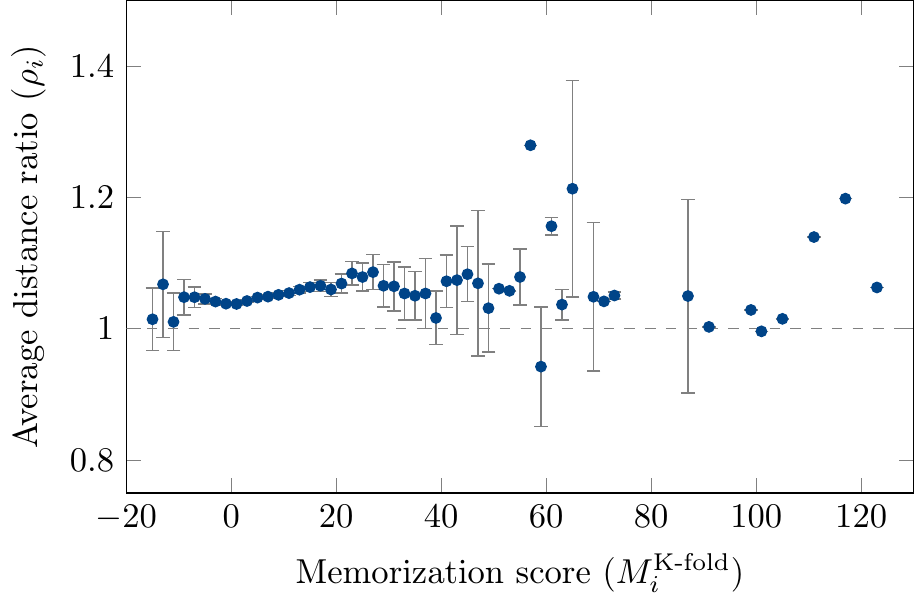}
		\caption{$M^{\text{K-fold}}_i$ vs. $\rho_i$ 
			\label{fig:nn_vs_mem_scatter_mnist_lr3}}
	\end{subfigure}
	\qquad
	\qquad
	\begin{subfigure}[b]{.40\textwidth}
		\includegraphics[height=40mm]{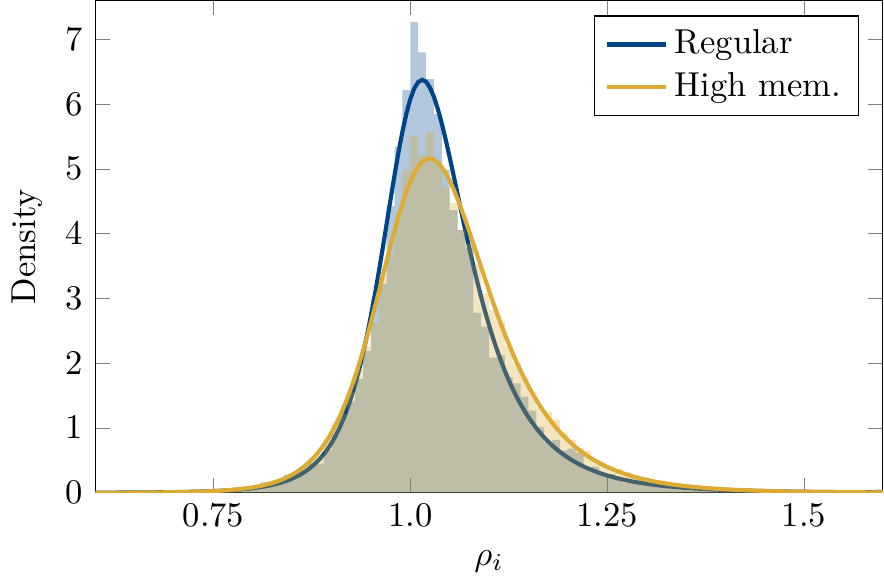}
		\caption{Distribution of $\rho_i$ 
			\label{fig:nn_vs_mem_hist_mnist_lr3}}
	\end{subfigure}
	\caption{Illustration of the nearest neighbor distance ratio in 
		(\ref{eq:nn_ratio}) compared to the memorization score for a 
		VAE trained on MNIST using a learning rate of $\eta = 
		10^{-3}$. We present the average of $\rho_i$ for bins of the 
		memorization score of width $2$, and show error bars 
		representing the confidence interval of the standard error of 
		the mean of distance ratio measurements in each bin.  The 
		horizontal axis in figure~(a) trims off one observation at 
		$M^{\text{K-fold}}_i \approx 210$ for clarity. Figure~(b) 
		shows the distribution of the distance ratio for observations 
		with a high memorization score (top 5\%) and the regular 
		ones.}%
	\label{fig:nn_vs_mem_mnist_lr3}
\end{figure}

\begin{figure}[h]
	\centering
	\captionsetup[subfigure]{justification=centering}%
	\begin{subfigure}[b]{.40\textwidth}
		\includegraphics[height=40mm]{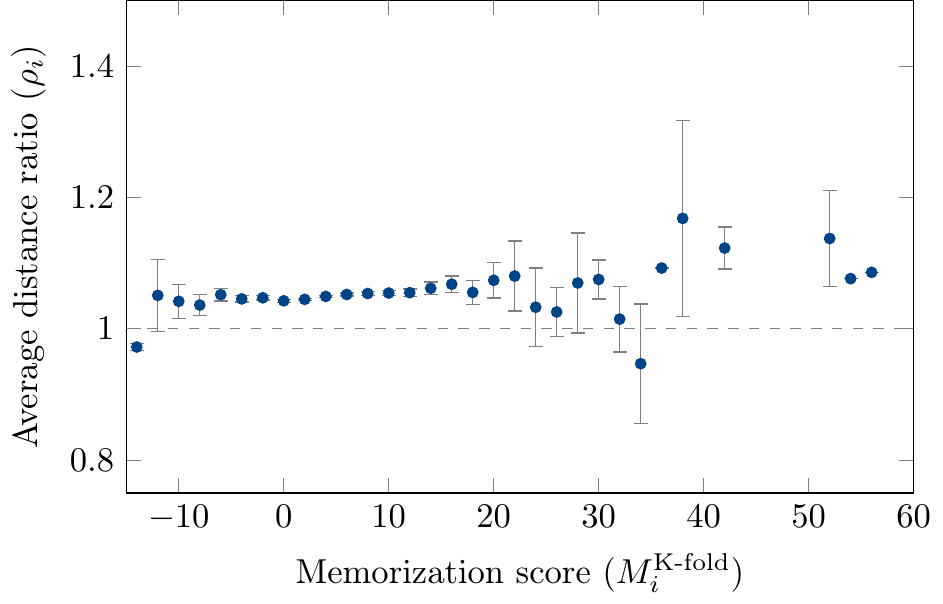}
		\caption{$M^{\text{K-fold}}_i$ vs. $\rho_i$ 
			\label{fig:nn_vs_mem_scatter_mnist_lr4}}
	\end{subfigure}
	\qquad
	\qquad
	\begin{subfigure}[b]{.40\textwidth}
		\includegraphics[height=40mm]{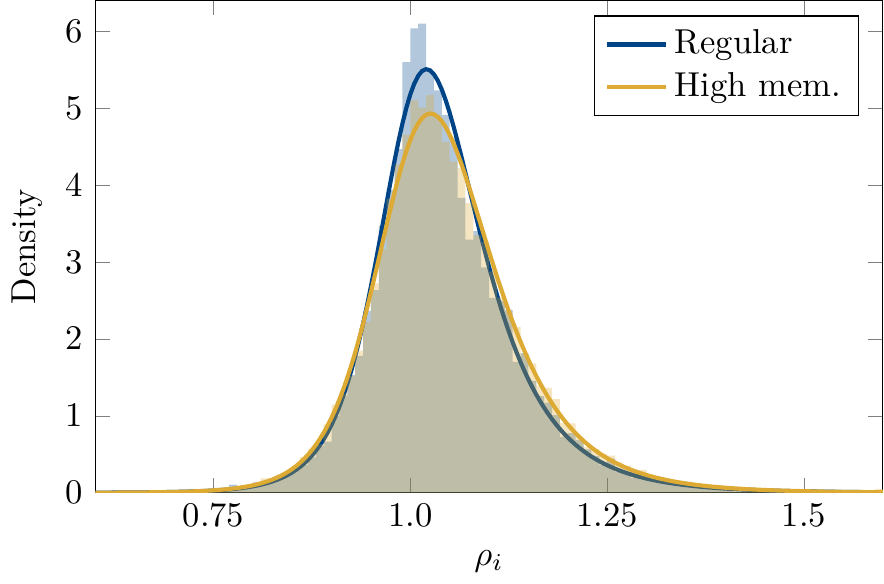}
		\caption{Distribution of $\rho_i$ 
			\label{fig:nn_vs_mem_hist_mnist_lr4}}
	\end{subfigure}
	\caption{Similar to Figure~\ref{fig:nn_vs_mem_mnist_lr3} but for the 
		VAE trained on MNIST using a learning rate of $\eta = 
		10^{-4}$. The horizontal axis in figure~(a) trims off one 
		observation at $M^{\text{K-fold}}_i \approx 80$ for clarity.}%
	\label{fig:nn_vs_mem_mnist_lr4}
\end{figure}
\end{document}